\documentclass{article}

\usepackage{tabularx} 
\usepackage{booktabs} 
\usepackage{graphicx} 
\usepackage{adjustbox} 
\PassOptionsToPackage{pagebackref=true}{hyperref}
\usepackage{PRIMEarxiv}
\usepackage{amsmath}
\usepackage[utf8]{inputenc} 
\usepackage[T1]{fontenc}   
\usepackage{hyperref}     
\usepackage{url}           
\usepackage{booktabs}       
\usepackage{amsfonts}       
\usepackage{nicefrac}      
\usepackage{microtype}      
\usepackage{lipsum}
\usepackage{fancyhdr}       
\usepackage{graphicx}       
\graphicspath{{media/}}     
\usepackage{subcaption}
\usepackage{url}

\usepackage{graphicx}
\usepackage{booktabs}
\usepackage[pagebackref=true]{hyperref}

\usepackage[accsupp]{axessibility}  

\usepackage{amsmath}
\usepackage{adjustbox}

\usepackage{algpseudocode}

\usepackage{marvosym} 
\usepackage{float}
\usepackage{stfloats} 
\usepackage{graphicx} 

\usepackage{lipsum}   
\usepackage{pdflscape}  

\usepackage[table]{xcolor}  
\usepackage{caption}        
\usepackage{subcaption} 
\usepackage{tikz}
\usetikzlibrary{shadows}

\usepackage{pgfplots}
\usepackage{siunitx}
\pgfplotsset{compat=1.17}
\usepackage{tikz}
\usepackage{pgfplots}
\usepackage{graphicx}
\usepackage{array}

\usepackage{pgfplots}

\usepackage{makecell}
\usepackage{diagbox}
\usepackage{appendix}

\usepackage{graphicx}
\usepackage{tabularx}      
\usepackage{arydshln}

\usepackage{colortbl}
\usepackage{booktabs}
\usepackage{multirow}      

\definecolor{best}{RGB}{204, 255, 204}     
\definecolor{secondbest}{RGB}{255, 235, 156}

\definecolor{mydarkblue}{rgb}{0,0.08,0.45}
\definecolor{mydarkred}{rgb}{0.6,0,0}
\definecolor{myblue}{HTML}{268BD2}
\definecolor{mygreen}{HTML}{658354}
\definecolor{results}{RGB}{220, 230, 240}

\usepackage[most]{tcolorbox}

\usepackage{algorithm}

\usepackage{orcidlink}

\title{SkyLink: A Large Vision-Language Model Driven Re-ranking Framework for Cross-View UAV geolocalization

}
\author{
  Bowen Liu\thanks{Equal contribution} $^1$, 
  Pengyue Jia\footnotemark[1] $^1$, 
  Wanyu Wang$^2$, 
  Derong Xu$^1$, 
  Jiawei Cheng$^1$, \\
  \textbf{Jiancheng Dong}$^1$, 
  \textbf{Xiao Han}$^3$, 
  \textbf{Zimo Zhao}$^1$, 
  \textbf{Chao Zhang}$^1$, 
  \textbf{Bowen Yu}$^1$, \\
  \textbf{Fangyu Hong}$^1$, 
  \textbf{Xiangyu Zhao}\thanks{Corresponding author} $^1$ \\
  \\
  $^1$Department of Data Science, City University of Hong Kong, Hong Kong \\
  $^2$Information Systems, City University of Hong Kong, Hong Kong \\
  $^3$College of Computer Science and Technology, Zhejiang University of Technology, Zhejiang \\
  \\
  \texttt{\{boweliu6-c, jia.pengyue\}@my.cityu.edu.hk} \\
  \quad \texttt{xianzhao@cityu.edu.hk}
}

\begin{document}
\maketitle

\begin{abstract}
Cross-view UAV geolocalization is fundamentally a challenging large-scale image retrieval task, aiming to determine the geographic coordinates of Unmanned Aerial Vehicle (UAV) queries by matching them against an extensive geo-tagged satellite image database. 
Most existing methods learn separate feature representations for each view and determine the final prediction using naive heuristics to assess feature similarity, thereby neglecting to model the crucial cross-view relationships. In this paper, we propose SkyLink, a novel plug-and-play ranking framework that pioneers joint relational modeling of inter-view relationships to enhance cross-view UAV geolocalization. 
SkyLink leverages a Large Vision-Language Model (LVLM) to model the intricate visual-semantic relationships between UAV and satellite views, facilitating effective cross-view matching. 
To further refine the learning process, we introduce a relational-aware loss. It leverages soft labels to provide a more nuanced supervision signal, mitigating the harsh penalty on near-positive pairs. This approach enhances both training stability and the model's discriminative capacity.
Extensive experiments conducted across multiple base retrieval architectures and benchmark datasets demonstrate that SkyLink significantly boosts the ranking effectiveness of existing models, consistently achieving superior performance in various challenging scenarios. 

\end{abstract}


\section{Introduction}
Cross-view geolocalization for unmanned aerial vehicles (UAVs) involves estimating the geographic position of a UAV-captured image by matching it to a reference database of geo-tagged overhead imagery, such as satellite views \cite{10506965}. This task has emerged as a critical alternative to satellite-based navigation systems (e.g., GPS \cite{Aughey2011}), which may fail in contested environments due to signal jamming or environmental interference. By exploiting visual cues from disparate viewpoints, cross-view geolocalization offers a robust solution in scenarios where conventional methods are unreliable, enabling applications in autonomous navigation \cite{1,2}, disaster response \cite{3,4}, precision agriculture \cite{5}, and urban planning \cite{6}. However, the task remains challenging due to the significant domain gap between UAV-oblique and satellite-orthographic images, including differences in scale, illumination, occlusion, and seasonal appearance \cite{durgam2024cross}. Addressing these discrepancies requires advanced feature matching and domain adaptation techniques. 

Recent progress \cite{10636268,10644040,liu2024segcn} in cross-view UAV geolocalization has primarily relied on dual-stream architectures, where separate encoders process UAV-oblique and satellite-orthographic images to extract view-specific features. During inference, these methods retrieve a set of candidates from the satellite image pool and select the one with the highest similarity score as the final location prediction.
While existing methods explore various backbone networks—including convolutional neural networks~\cite{Deuser_2023_ICCV,rodrigues2023semgeo,durgam2024cross}, Transformers~\cite{ji2025game4loc,liu2024segcn}, and state space models~\cite{huangvimgeo}—to improve representation learning, they all eventually fall back to naive heuristics (e.g., cosine similarity \cite{Xia2015}) for scoring cross-view relevance. This heuristic ranking strategy ignores the intricate semantic and spatial relationships between UAV and satellite views, which are essential for closing the large domain gap and improving candidate selection accuracy.
Moreover, current contrastive learning based training objectives, including InfoNCE~\cite{oord2018representation} and triplet loss~\cite{schroff2015facenet}, impose uniform penalties on all negative pairs, regardless of their visual or geographic proximity to the ground truth. This overly rigid supervision can hinder convergence and reduce discriminative ability in challenging scenarios, ultimately constraining model robustness and generalization.

To address the aforementioned limitations, we introduce SkyLink, a novel ranking framework that joins relational modeling for cross-view UAV geolocalization. 
Unlike previous methods that model UAV and satellite views separately, SkyLink leverages an LVLM to jointly model the cross-view interaction in a unified representation space and directly outputs a scalar relevance score for each candidate.

This unified modeling enables the network to capture semantic correspondences and spatial alignments across views, which are often missed by independent encoders followed by naive similarity measures.
Furthermore, inspired by recent advances in graded similarity supervision~\cite{Leyva-Vallina_2023_CVPR, 10611288}, we propose a relational-aware loss that incorporates soft labels to provide nuanced supervision, assigning graduated penalties to near-positive pairs. This design stabilizes the training process and ultimately leads to improved performance.

Addressing the lack of datasets dedicated to training ranking models, we curated SkyRank from open-source benchmarks. We first employ an existing cross-view geolocalization model as a retriever to generate candidate sets from a large satellite image pool for each UAV query image. The ground truth is either present in the retrieved candidates or explicitly inserted, ensuring that the model learns to identify the correct match from among them. While this preprocessing introduces a specific inductive bias, this intentional distribution shift aligns the training distribution with that of the inference phase. This alignment compels the model to focus on distinguishing hard samples that share global similarities yet differ. To support future research in cross-view geolocalization and broader vision-language tasks, we will release SkyRank. 

This paper validates the effectiveness of the proposed method by evaluating SkyLink under task settings, based on combinations of three retrievers (Sample4geo~\cite{Deuser_2023_ICCV}, SDPL~\cite{10587023}, MCCG~\cite{10185134}) and two benchmark datasets (University-1652~\cite{DBLP:conf/mm/ZhengWY20}, SUES-200~\cite{DBLP:journals/tcsv/ZhuYYWYH23}). The results demonstrate that the retriever combinations with SkyLink as a plugin achieve comprehensive improvements in metrics such as recall rate and average precision. Furthermore, extensive ablation studies confirm the effectiveness of each proposed component, while hyperparameter sensitivity analysis provides profound insights into model stability. Our contributions can be summarized as follows:

\begin{itemize}
    \item We propose SkyLink, a novel plug-and-play LVLM-based framework for cross-view UAV geolocalization that jointly models interactions across UAV and satellite views within a unified representation space. This joint modeling improves relevance scoring via fine-grained cross-view correspondences, overcoming limitations of prior heuristic measures in the literature.
    \item We construct and release SkyRank datasets for cross-view geolocalization. SkyRank is curated to support the training of ranking-based methods. By releasing this derived benchmark, we aim to provide a useful resource for further exploration in cross-view geolocalization and related fields.
    \item Extensive experiments are conducted within a single task setting, involving combinations of three retrievers and two benchmark datasets. The results demonstrate that SkyLink comprehensively enhances the retrieval performance of the retrievers.

\end{itemize}

\section{Methodology}

\begin{figure*}[t]
\centering
\includegraphics[width=1\textwidth,scale=0.1]{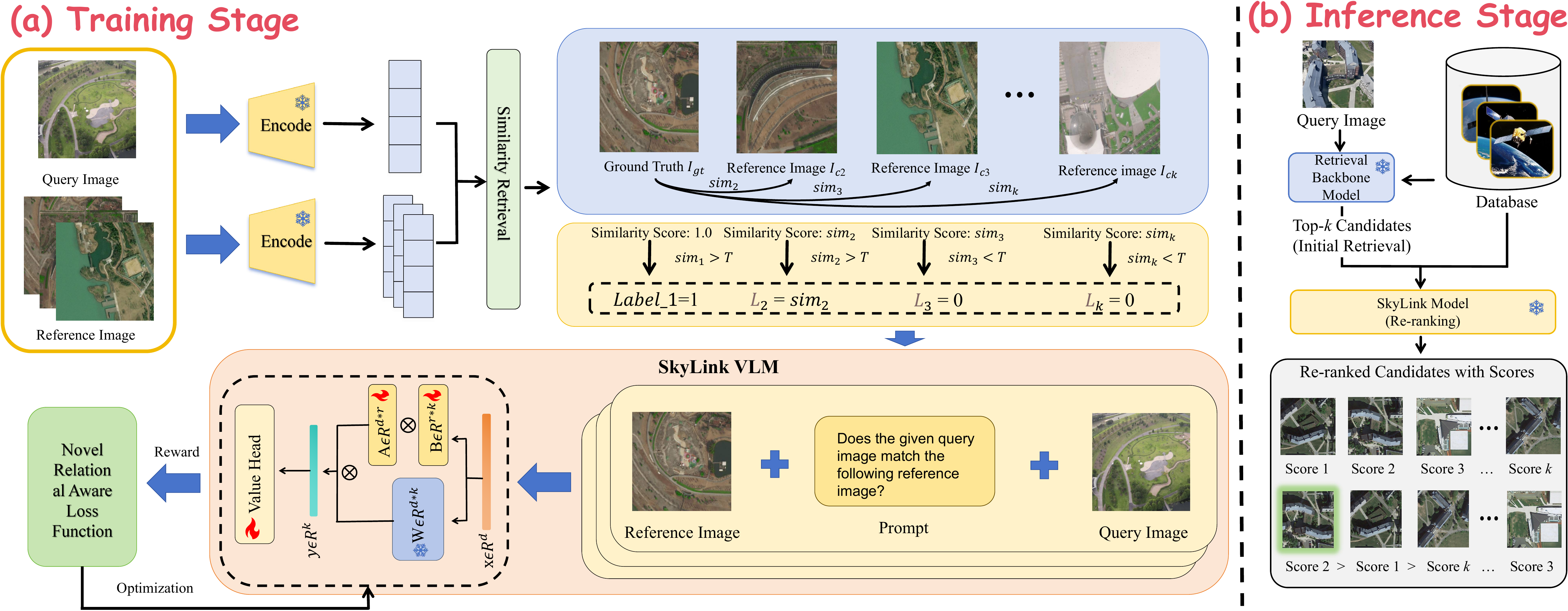}
    \caption{The Framework of SkyLink. During the training phase, the model takes the retrieval results generated by the retriever on the training set as input. In the inference phase, the model receives the retrieval results output by the retriever on the test set, and optimizes them through re-ranking.}
    \label{figure.1}
\end{figure*}

In this section, we introduce SkyLink, our proposed ranking framework for cross-view UAV geolocalization that models the interaction between different views. As illustrated in Figure~\ref{figure.1}, the overall framework of SkyLink comprises two main stages: training stage and inference stage.

\subsection{Dataset Curation}

Since no off-the-shelf dataset is available for training ranking models in cross-view geolocalization, we construct new datasets from existing public benchmarks to support the training of SkyLink. Given a standard cross-view geo-localization dataset, we denote a query UAV image as $I_q$, its corresponding ground truth satellite image as $I_{gt}$, and the entire pool of reference satellite images as $\mathcal{R} = \{I_{r_i}\}_{i=1}^N$.

To generate a set of candidate images for each query, we first employ a pre-trained cross-view geolocalization model as a retriever. We build a feature gallery $\mathcal{V_R}$ by encoding every reference image $I_{r_i} \in \mathcal{R}$ using the retriever's reference encoder $\mathcal{E}_r(\cdot)$, such that $\mathcal{V_R} = \{\mathbf{v}_{r_i} | \mathbf{v}_{r_i} = \mathcal{E}_r(I_{r_i})\}$. For each query image $I_q$, we extract its feature vector $\mathbf{v}_q = \mathcal{E}_q(I_q)$. We then retrieve the top-$m$ most similar candidates $\mathcal{C}_q$ from the reference pool by matching all reference features based on their cosine similarity to the query feature:
\begin{equation}
    \mathcal{C}_q 
    = \left\{ 
        I_{r_i} \;\middle|\; 
        I_{r_i} \in \operatorname*{top\text{-}m}_{I_{r_j} \in \mathcal{R}}
        \mathrm{sim}\!\left(\mathbf{v}_q, \mathbf{v}_{r_j}\right) 
    \right\},
\end{equation}
where we define the similarity function as the cosine similarity $\text{sim}(\mathbf{v}_q, \mathbf{v}_{r_i}) = 
\mathbf{v}_q^\top \mathbf{v}_{r_i} / \|\mathbf{v}_q\|\;\|\mathbf{v}_{r_i}\|$.

The retrieved set $\mathcal{C}_q$ contains hard negative samples but may or may not include the ground truth $I_{gt}$. To handle these scenarios, we curate the candidate set $\mathcal{C}_q$ as follows:
\begin{equation}
\mathcal{C}_q^{\prime} = 
\begin{cases}
    \mathcal{C}_q \setminus \{I_{gt}\}, & \text{if } I_{gt} \in \mathcal{C}_q \\
    \mathcal{C}_q \setminus \{I_{r_m}\}, & \text{if } I_{gt} \notin \mathcal{C}_q
\end{cases},
\end{equation}
where $I_{r_m}$ is the least similar (the $m$-th) candidate in $\mathcal{C}_q$. This process ensures that $\mathcal{C}_q^{\prime}$ consistently contains exactly $m-1$ candidates for every query. Finally, each sample in our curated dataset consists of a tuple containing the query image $I_q$, the set of $m-1$ candidates $\mathcal{C}_q^{\prime}$, and the ground truth image $I_{gt}$.

\subsection{SkyLink}

Existing approaches \cite{Deuser_2023_ICCV,rodrigues2023semgeo,durgam2024cross} typically adopt a dual-stream architecture, where images from different views are encoded separately and then projected into a shared representation space. The final candidates are selected based on the similarity between these representations. While substantial effort has been devoted to improving feature extraction within each stream, much less attention has been paid to modeling the interactions across different views. The commonly used naive heuristics (e.g., cosine similarity) often fail to capture the complex semantic and spatial relationships between views, resulting in suboptimal matching performance. To address this limitation, we propose SkyLink, a novel ranking framework that models cross-view interactions. Specifically, images from different views are first combined using a predefined prompt template to form a unified model input. SkyLink then leverages a powerful LVLM backbone to capture the intricate relationships across views. Finally, a value head is appended to the last layer of the LVLM to produce a scalar relevance score, which reflects the matching degree between the query and candidate images.

\paragraph{Cross-View Prompt Template.}

To leverage LVLMs, we integrate multi-view images into a unified input. Given a sample $(I_q, C_q^{\prime}, I_{gt})$, we employ a hard negative mining strategy rather than random sampling. Specifically, we select the top-$(k-1)$ candidates from $C_q^{\prime}$ based on retriever similarity. These 'hard negatives'—visually similar to $I_q$ yet geographically distinct—are combined with $I_{gt}$ to form the candidate set. This design compels the model to distinguish among highly ambiguous samples. For each candidate in this set, together with the query image $I_q$, we construct an input using the following prompt template: <query image> \textit{ Does the given query image match the following reference image?} <candidate image>

As a result, each training sample yields $k$ prompt inputs. Formally, this process can be expressed as:
\begin{equation}
\mathcal{P} = \big\{\, \text{Prompt}(I_q, I_{c_j}) \;\big|\; I_{c_j} \in \{I_{gt}\} \cup \text{top-}(k-1)(C_q^{\prime}) \big\},
\end{equation}
where $\text{Prompt}(\cdot,\cdot)$ denotes the application of the cross-view template to a query--candidate pair.

\paragraph{Model Architecture.}
The constructed prompt set $\mathcal{P}$ is fed into a pretrained LVLM backbone to capture the semantic and spatial interactions between the query and candidate images.
To adapt the model to the cross-view matching task, we insert Low-Rank Adaptation (LoRA)~\cite{hu2022lora} modules into each transformer layer of the LVLM.
These lightweight adapters enable efficient fine-tuning and improve the model's expressiveness over multi-view inputs.
Specifically, for a given transformer layer $l$, we apply LoRA to the projection matrix $\mathbf{W}_l$ as follows:
\begin{equation}
    \tilde{\mathbf{W}}_l = \mathbf{W}_l + \Delta \mathbf{W}_l = \mathbf{W}_l + \mathbf{A}_l \mathbf{B}_l,
\end{equation}
where $\tilde{\mathbf{W}}_l \in \mathbb{R}^{d \times d}$ is the updated weight matrix, $\mathbf{W}_l \in \mathbb{R}^{d \times d}$ is the original frozen weight matrix from the LVLM, and $\mathbf{A}_l \in \mathbb{R}^{d \times r}$, $\mathbf{B}_l \in \mathbb{R}^{r \times d}$ are the trainable low-rank matrices with rank $r \ll d$. During training, only $\mathbf{A}_l$ and $\mathbf{B}_l$ are updated, enabling efficient adaptation to the cross-view matching task with minimal additional parameters.

We extract the hidden state of the last token from the final layer of the LVLM as a compact summary of the cross-view input. To produce a scalar relevance score, we apply a lightweight value head—implemented as a single bias-free linear layer—on top of this representation:
\begin{equation}
    S = \mathbf{w}^\top \mathbf{h}_{\text{[last]}}, \text{where} \ \mathbf{h}_{\text{[last]}}=\text{LVLM}(\mathcal{P})_{\text{[-1]}},
\end{equation}
where $S \in \mathbb{R}^{k}$ is the output score vector for the $k$ candidates, $\mathbf{w}$ is the weight vector of the value head, and
$\mathbf{h}_{\text{[last]}}$ denotes the hidden state of the last token from the final layer of the LVLM. The resulting score reflects the predicted matching degree between the query and candidate image pair.

\subsection{Relational-aware Optimization}

Previous works ~\cite{mi2024congeo,Deuser_2023_ICCV,rodrigues2023semgeo,durgam2024cross,avola2024uav}, constrained by their dual-stream architectures, typically rely on InfoNCE or triplet loss to optimize separate encoders for each view. However, these objectives are not only illsuited for training a ranking framework like SkyLink, but also share a critical limitation: they penalize all negative samples uniformly, failing to account for their visual or geographic proximity to the ground truth. Such rigid supervision can impair model convergence and diminish its discriminative power in challenging scenarios. To address these limitations, we propose a relational-aware loss function to train SkyLink. This loss is designed to guide the model in selecting the correct match from a list of candidates while providing soft labels to near-positive examples, which significantly enhances training stability.
\paragraph{Training Objective.}
Given the output score vector $S \in \mathbb{R}^k$ from the LVLM and the candidate set $C_q^{\prime} = \{I_{c_1}, I_{c_2}, \dots, I_{c_k}\}$, along with the ground-truth reference image $I_{gt}$ and the retriever's reference encoder $\mathcal{E}_r$, we construct soft labels based on the similarity between each candidate and the ground truth in the feature space. Specifically, for each candidate image $I_{c_j} \in C_q^{\prime}$, we compute its cosine similarity with the ground-truth image:
\begin{equation}
    \text{sim}_j = \frac{\mathcal{E}_r(I_{c_j})^\top \mathcal{E}_r(I_{gt})}{\|\mathcal{E}_r(I_{c_j})\| \cdot \|\mathcal{E}_r(I_{gt})\|},
\end{equation}
It is worth noting that, due to the way we construct the dataset, the candidate set $C_q^{\prime}$ is guaranteed to contain the ground truth image $I_{gt}$, whose similarity score is always 1.0.
We then define a soft label vector $L \in \mathbb{R}^k$ as follows:
\begin{equation}
    L_j =
    \begin{cases}
        \text{sim}_j, & \text{if } \text{sim}_j > $\textit{T}$, \\
        0, & \text{otherwise},
    \end{cases}
\end{equation}
where $T$ is a similarity threshold that filters out unrelated candidates.

Furthermore, due to the lack of sufficient metadata in the target datasets, precise computation of geometric intersection and distance metrics is infeasible. In this setting, the feature-space similarity derived from a well-calibrated cross-view retriever, coupled with our carefully curated re-ranking dataset, effectively mitigates performance variability induced by discrepancies across different retriever architectures.

We train the model using the binary cross-entropy loss with logits. The loss is computed between the predicted scores $S$ and the soft labels $L$ as follows:
\begin{equation}
    \mathcal{L} = -\frac{1}{k} \sum_{j=1}^k \left[ L_j \cdot \log \sigma(S_j) + (1 - L_j) \cdot \log (1 - \sigma(S_j)) \right],
\end{equation}
where $\sigma(\cdot)$ denotes the sigmoid function. This formulation allows the model to learn fine-grained cross-view relevance supervision based on feature-space proximity, rather than relying on binary labels alone.

\subsection{Inference}

\paragraph{Inference.}
During inference, we first retrieve a candidate set $C_q$ from the reference image pool based on the query image $I_q$. Each query--candidate pair $(I_q, I_{c_j})$ is then used to construct an input prompt according to the predefined template. These prompts are fed into the SkyLink model to compute a matching score for each pair:
\begin{equation}
    s_j = \text{SkyLink}(\text{Prompt}(I_q, I_{c_j})), \quad \text{for } I_{c_j} \in C_q.
\end{equation}
Then, we select the candidate with the highest score as the predicted match:
\begin{equation}
    \hat{I} = \underset{I_{c_j} \in C_q}{\arg\max} \; s_j.
\end{equation}
This inference procedure enables accurate cross-view geolocalization by ranking the candidates based on their semantic and spatial relevance to the query.

\section{Experiments}

\subsection{Experimental Setup}

\paragraph{Datasets and Evaluation Metrics.}

We conduct experiments on two datasets of cross-view geolocalization tasks. The experiment is an image-based UAV geolocalization task, for which we use two widely adopted benchmark datasets: University-1652~\cite{DBLP:conf/mm/ZhengWY20} and SUES-200~\cite{DBLP:journals/tcsv/ZhuYYWYH23}. Following the original dataset protocols, we evaluate performance using Recall and Average Precision (AP). For specific details regarding the datasets and evaluation metrics, please refer to Appendix~\ref{a1}.

\paragraph{SkyRank Dataset.}
In this study, we curated a derived dataset tailored for training candidate ranking models in cross-view geolocalization. Specifically, for each UAV query image, we retrieved a set of candidate samples based on the embedding similarity derived from the training set with a retriever (cross-view geolocalization model). Subsequently, we ensured the presence of the ground-truth match in each candidate set by either retaining it if already retrieved or explicitly inserting it during the subsequent data cleaning process. For each combination of retrievers, we constructed a corresponding re-ranking dataset. The purpose of releasing this dataset is to provide support for the development of cross-view geolocalization and related research fields, such as UAV navigation, information retrieval, and large vision-language models (LVLMs). For example entries of the dataset, please refer to Appendix~\ref{a2}.

\paragraph{Implementation Details.}

Our retriever is built upon three state-of-the-art cross-view geo-localization retrievers: SDPL~\cite{10587023}, Sample4Geo~\cite{Deuser_2023_ICCV} and MCCG~\cite{10185134}. (For more details of retrievers, refer to Appendix ~\ref{a3}) To ensure a strictly fair comparison, all retrievers are either initialized with official pre-trained weights that already encompass our target datasets, or undergo pre-training on the same dataset using their officially recommended configurations. During the training phase, 20 candidate samples are retrieved from the database for each query sample. Among these, the top-13 samples are used as retrieval candidate samples. In SkyLink, Qwen2-VL-7b-Instruct is adopted as the backbone network of the Large Vision-Language Model (LVLM). For LoRA (Low-Rank Adaptation) fine-tuning, the target modules include \texttt{q\_proj}, \texttt{k\_proj}, \texttt{v\_proj}, and visual LoRA. SkyLink is fine-tuned using the AdamW \cite{loshchilov2019decoupledweightdecayregularization} optimizer. All experiments are conducted based on the PyTorch framework and completed on 2 NVIDIA A100 GPUs (80GB). In the inference phase, the retrievers are used to generate candidate samples for the LVLM. The  similarity threshold \(T\) is set to 0.9; Throughout the entire inference phase, re-ranking is only performed on the top-10 reference images retrieved by the retrievers. Additional details on the training environment and runtime are provided in Appendix~\ref{a4}.

\subsection{Main Results}
To systematically evaluate the generality and incremental optimization potential of the SkyLink model as a plug-and-play re-ranking module, this study conducted comprehensive and rigorous comparative experiments on multiple representative geospatial image retrieval benchmark datasets with different base retrievers. As shown in Table~\ref{tab:university1652_r1_ap}, after re-ranking by the SkyLink model, all base retrievers achieved significant performance improvements on datasets such as University-1652 and SUES-200. For example, in the drone-to-street (D2S) task, the model exhibited particularly prominent enhancement effects on systems with SDPL and MCCG as base retrievers — it improved the R@1 metric of SDPL by 8.7\% in the University-1652 dataset and further boosted the recall rate of MCCG by 17.17\% in the 150m low-altitude drone retrieval task on the SUES-200 dataset, fully demonstrating its excellent performance optimization potential. Notably, although MCCG~\cite{10185134} and Sample4Geo~\cite{Deuser_2023_ICCV} among the retrievers have shown relatively superior retrieval performance, the inherent modal information isolation of the traditional dual-stream architecture and the intrinsic limitation of the InfoNCE loss function in distinguishing ambiguous samples restrict their performance upper bound. In contrast, the SkyLink model breaks through the above bottlenecks through two core innovations: firstly, it abandons the modal isolation constraint of the traditional dual-stream architecture and constructs a cross-modal image-text interaction mechanism; secondly, it proposes a dynamic relation-aware loss function, which achieves accurate distinction of ambiguous samples via a soft-label modeling strategy, thereby significantly improving the retrieval accuracy in the re-ranking stage. In summary, our method yields substantial improvements in retrieval performance of retrievers across all datasets, fully verifying its effectiveness and superiority.

\begin{table*}[t]
\centering
\footnotesize
\caption{Comprehensive Performance Comparison on University-1652 and SUES-200 Datasets Based on Various Retrievers.}
\label{tab:university1652_r1_ap}
\setlength{\tabcolsep}{3pt}

\definecolor{scigreen}{RGB}{0, 100, 0} 
\newcommand{\imp}[1]{\textcolor{scigreen}{\scriptsize $\uparrow$#1}}

\newcommand{\tc}[1]{\makebox[3.2em]{\textbf{#1}}}

\resizebox{\textwidth}{!}{%
\begin{tabular}{@{}l *{6}{c} *{8}{c} @{}}
\toprule
\multirow{3}{*}{\textbf{Method}} 
& \multicolumn{6}{c}{\textbf{University-1652}} 
& \multicolumn{8}{c}{\textbf{SUES-200 Dataset}} \\
\cmidrule(lr){2-7} \cmidrule(l){8-15}
& \multicolumn{3}{c}{\textbf{D2S}} & \multicolumn{3}{c}{\textbf{S2D}} 
& \multicolumn{2}{c}{\textbf{150m}} & \multicolumn{2}{c}{\textbf{200m}} 
& \multicolumn{2}{c}{\textbf{250m}} & \multicolumn{2}{c}{\textbf{300m}} \\
\cmidrule(lr){2-4} \cmidrule(lr){5-7} \cmidrule(lr){8-9} \cmidrule(lr){10-11} \cmidrule(lr){12-13} \cmidrule(lr){14-15}

& \tc{R@1} & \tc{R@5} & \tc{AP} 
& \tc{R@1} & \tc{R@5} & \tc{AP}
& \tc{R@1} & \tc{AP} & \tc{R@1} & \tc{AP}
& \tc{R@1} & \tc{AP} & \tc{R@1} & \tc{AP} \\
\midrule

SDPL \cite{10587023} 
& 85.17 & 94.96 & 89.61 & 89.30 & 92.87 & 89.49 
& 79.65 & 83.03 & 86.48 & 89.08 & 93.33 & 94.70 & 93.15 & 94.53 \\

\hspace{1em}+SkyLink (Ours)
& 93.87\cellcolor{best} & 96.57\cellcolor{best} & 95.13\cellcolor{best} & 91.87\cellcolor{best} & 93.44\cellcolor{best} & 91.57\cellcolor{best}
& 90.97\cellcolor{best} & 93.15\cellcolor{best} & 94.13\cellcolor{best} & 96.39\cellcolor{best} & 94.57\cellcolor{best} & 96.76\cellcolor{best} & 94.65\cellcolor{best} & 96.44\cellcolor{best} \\

\multicolumn{1}{r}{\textit{Improvement}$^1$} & 
\imp{8.70} & \imp{1.61} & \imp{5.52} & \imp{2.57} & \imp{0.57} & \imp{2.08}
& \imp{11.32} & \imp{10.12} & \imp{7.65} & \imp{7.31} & \imp{1.24} & \imp{2.06} & \imp{1.50} & \imp{1.91} \\

\addlinespace

MCCG \cite{10185134} 
& 89.58 & 96.55 & 92.72 & 93.44 & 96.29 & 93.93
& 76.20 & 84.49 & 89.42 & 93.54 & 94.45 & 96.55 & 96.75 & 98.25 \\

\hspace{1em}+SkyLink (Ours)
& 94.93\cellcolor{best} & 97.79\cellcolor{best} & 96.27\cellcolor{best} & 93.58\cellcolor{best} & 97.15\cellcolor{best} & 94.49\cellcolor{best}
& 93.37\cellcolor{best} & 95.03\cellcolor{best} & 95.67\cellcolor{best} & 97.36\cellcolor{best} & 96.50\cellcolor{best} & 97.78\cellcolor{best} & 97.52\cellcolor{best} & 98.52\cellcolor{best} \\

\multicolumn{1}{r}{\textit{Improvement}$^1$} &
\imp{5.35} & \imp{1.24} & \imp{3.55} & \imp{0.14} & \imp{0.86} & \imp{0.56}
& \imp{17.17} & \imp{10.54} & \imp{6.25} & \imp{3.82} & \imp{2.05} & \imp{1.23} & \imp{0.77} & \imp{0.27} \\

\addlinespace

Sample4geo \cite{Deuser_2023_ICCV} 
& 92.05 & 97.55 & 94.49 & 94.58 & 96.72 & 94.82
& 92.87 & 95.58 & 94.47 & 96.79 & 96.83 & 98.17 & 97.40 & 98.16 \\

\hspace{1em}+SkyLink (Ours)
& 95.28\cellcolor{best} & 98.05\cellcolor{best} & 96.56\cellcolor{best} & 94.58\cellcolor{best} & 96.72\cellcolor{best} & 94.82\cellcolor{best}
& 95.77\cellcolor{best} & 97.51\cellcolor{best} & 96.45\cellcolor{best} & 98.02\cellcolor{best} & 97.75\cellcolor{best} & 98.70\cellcolor{best} & 97.62\cellcolor{best} & 98.76\cellcolor{best} \\

\multicolumn{1}{r}{\textit{Improvement}$^1$} &
\imp{3.23} & \imp{0.50} & \imp{2.07} & \imp{0.00} & \imp{0.00} & \imp{0.00}
& \imp{2.90} & \imp{1.93} & \imp{1.98} & \imp{1.23} & \imp{0.92} & \imp{0.53} & \imp{0.22} & \imp{0.60} \\

\bottomrule
\multicolumn{15}{l}{\scriptsize $^1$Improvement in model performance and baseline comparison.}
\end{tabular}%
}
\end{table*}

\subsection{Hyperparameter Analysis}

To gain an in-depth understanding of the impact of key hyperparameters, we conducted systematic ablation experiments following the experimental method of controlling variables. The hyperparameters under consideration include: (1) The number of retrieval candidates \( G_z \) in the training phase; (2) The number of retrieval candidates (top-\textit{k}) in the inference phase; (3) The similarity threshold  \( T \) for soft classification. Unless otherwise specified, the default parameter values are set as follows: \( |G_z| = 7 \), \( \text{top-\textit{k}} = 10 \), and \( |T| = 0.8 \) in this section. All experiments were implemented using Sample4geo as the retriever, with the University-1652 dataset serving as the experimental benchmark. Due to hardware constraints, the maximum value of \( |G_z| \) was limited to 13.

\begin{figure}[t]
\centering
\includegraphics[width=\textwidth,scale=1]{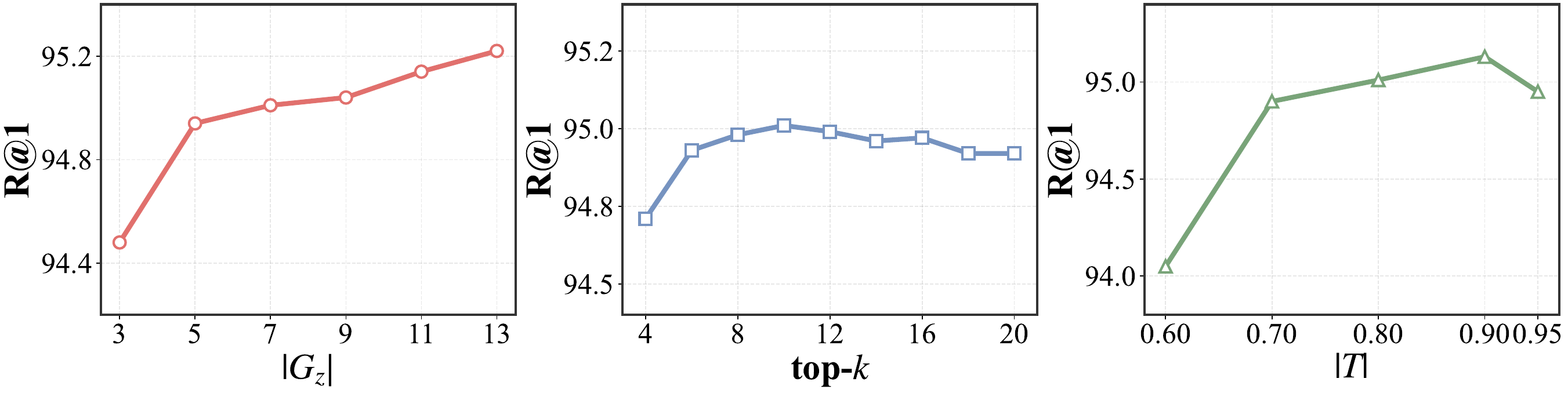}
    \caption{Hyperparameter Analysis on University-1652.}
    \label{5}
\end{figure}

We conduct a systematic analysis of the experimental results presented in Figure~\ref{5}. 

Regarding the candidate size \( |G_z| \) in the training phase, the model performance exhibits a continuous upward trend as \( |G_z| \) increases. This phenomenon can be attributed to the following: a moderate expansion of the candidate pool elevates the task complexity during the training process—by introducing more diverse reference samples, the model gains access to richer supervisory signals. These supervisory signals facilitate the model in learning fine-grained feature representations and discriminative decision boundaries, thereby enhancing its generalization ability on the target task.  
For the retrieval candidate size \( \text{top-\textit{k}} \) in the inference phase, the model performance shows a 'first rising and then stabilizing' trajectory. In the initial stage of \( \text{top-\textit{k}} \) increase, a larger candidate pool directly improves the probability of including the ground-truth candidate in the retrieval results, thus driving the continuous improvement of evaluation metrics. However, when \( \text{top-\textit{k}} \) exceeds a specific threshold, the expanded candidate pool inevitably introduces a large number of noisy samples (i.e., irrelevant candidates with low feature similarity to the query sample). Notably, even in the presence of such noise, the model performance remains stable, which strongly demonstrates that the proposed model possesses robust anti-noise capability. This robustness stems from the model’s effective feature filtering mechanism: during the re-ranking process, the model can prioritize the identification of high-confidence relevant candidates while suppressing the interference of noisy samples, ultimately maintaining reliable inference performance.

For the similarity threshold (|$T$|), the model performance exhibits a significant 'first increasing and then decreasing' trend, with its optimal performance achieved when $|T| = 0.9$. This phenomenon reveals that applying an appropriate penalty constraint on sample labels can effectively guide the model to explore the latent distribution patterns and feature correlations of similar samples in the candidate pool, thereby enhancing the model's ability to discriminate samples with subtle differences. This result demonstrates that by dynamically adjusting the constraint strength on sample similarity, a potential solution with both interpretability and practicality can be provided for the academic challenge of fine-grained distinction between positive samples and semi-positive samples. Notably, the selection of the hyperparameter |$T$| exhibits flexibility and can be adaptively adjusted based on the structural distribution of specific datasets. For further details on hyperparameter |$T$|, please refer to Appendix~\ref{a5}

\begin{table}[t]
  \caption{Ablation Study and Parameter Scale Analysis on University-1652.}
  \vspace{0.1cm} 
  \centering
  \begin{minipage}{0.48\linewidth}
    \centering
    \subcaption{Ablation Study.}
    \label{1}
    \resizebox{\linewidth}{!}{
      \begin{tabular}{@{}lccc@{}}
        \toprule
        Method & R@1$\uparrow$ & R@5$\uparrow$ & AP$\uparrow$ \\
        \midrule
        Ours                     & \textbf{95.28} & \underline{98.05} & \textbf{96.56} \\
        w/o soft positive label  & \underline{95.13} & 98.03          & \underline{96.47} \\
        w/o similarity threshold & 94.07          & \textbf{98.08} & 95.94          \\
        w/o SkyLink              & 92.05          & 97.55          & 94.49          \\
        \bottomrule
      \end{tabular}%
    }
  \end{minipage}\hfill
  \begin{minipage}{0.50\linewidth} 
    \centering
    \subcaption{Parameter Scale Analysis.}
    \label{2}
    \resizebox{\linewidth}{!}{
      \begin{tabular}{@{}lccc@{}}
        \toprule
        Backbone model & R@1$\uparrow$ & R@5$\uparrow$ & AP$\uparrow$ \\
        \midrule
        llava-onevision-0.5B  & 90.61 & 97.94  & 93.92 \\
        Qwen2-VL-2B-Instruct  & 91.18 & 97.96  & 94.25 \\
        Qwen2-VL-7B-Instruct  & 95.28 & 98.05  & 96.56 \\
        Improvement$^1$       & $\uparrow$ 4.67 & $\uparrow$ 0.11 & $\uparrow$ 2.64 \\
        \bottomrule
        \multicolumn{4}{l}{\scriptsize $^1$Maximum performance gain among models.}
        \vspace{-1.2em}
      \end{tabular}%
    }
  \end{minipage}
  \vskip -0.1in
\end{table}

\subsection{Ablation Study}

To deeply analyze the functional contributions and action mechanisms of each core component in the model, this study conducted ablation experiments using the systematic component removal method. 

The specific experimental settings of ablation study are defined as follows:
\begin{itemize}
    \item \textbf{w/o soft positive label}: Refers to the adoption of a hard label encoding mechanism in the model training process. When the cosine similarity between reference samples in the training set exceeds a preset similarity threshold, the label of the sample pair is directly hard-coded as 1, without introducing an elastic space for soft label classification.
    \item \textbf{w/o similarity threshold}: Refers to the exclusion of a penalty threshold mechanism based on cosine similarity during model training, i.e., no soft label classification based on similarity is performed for sample labels, and only a unified label definition method is adopted.

     \item \textbf{w/o SkyLink}: Refers to the complete removal of the SkyLink model, retaining only the baseline retrieval capability of the retriever.
\end{itemize}

Table~\ref{1}  presents the ablation experiment results of the proposed model on the University-1652 dataset. Based on the data in the table, the following key conclusions can be drawn:

\begin{enumerate}
    \item All core components in the model framework make significant positive contributions to the final performance of the model. This result directly verifies the rationality and effectiveness of the model design scheme proposed in this study. 
    
    \item A performance comparison between the complete model and the variant model with the soft encoding strategy removed shows that: within the reference sample range of the training set, the category discrimination ability of the variant model for ambiguous samples, positive samples, and semi-positive samples is significantly lower than that of the complete model. This phenomenon indicates that the flexible label encoding strategy can effectively enhance the model's ability to identify and distinguish ambiguous samples during the training phase, while providing more flexible technical support for exploring the potential correlations within the reference set and between reference samples.
    
    \item After removing the similarity threshold mechanism from the model, the experiment observed a phenomenon of severe fluctuations in the loss value during the entire training process of the model.  Please refer to Appendix~\ref{a6} for details on the further analysis.

\end{enumerate}

\subsection{Efficiency Analysis}

\begin{figure}[t]
    \centering
    
    \begin{subfigure}[b]{0.48\textwidth}
        \centering
       
        \includegraphics[width=0.85\linewidth]{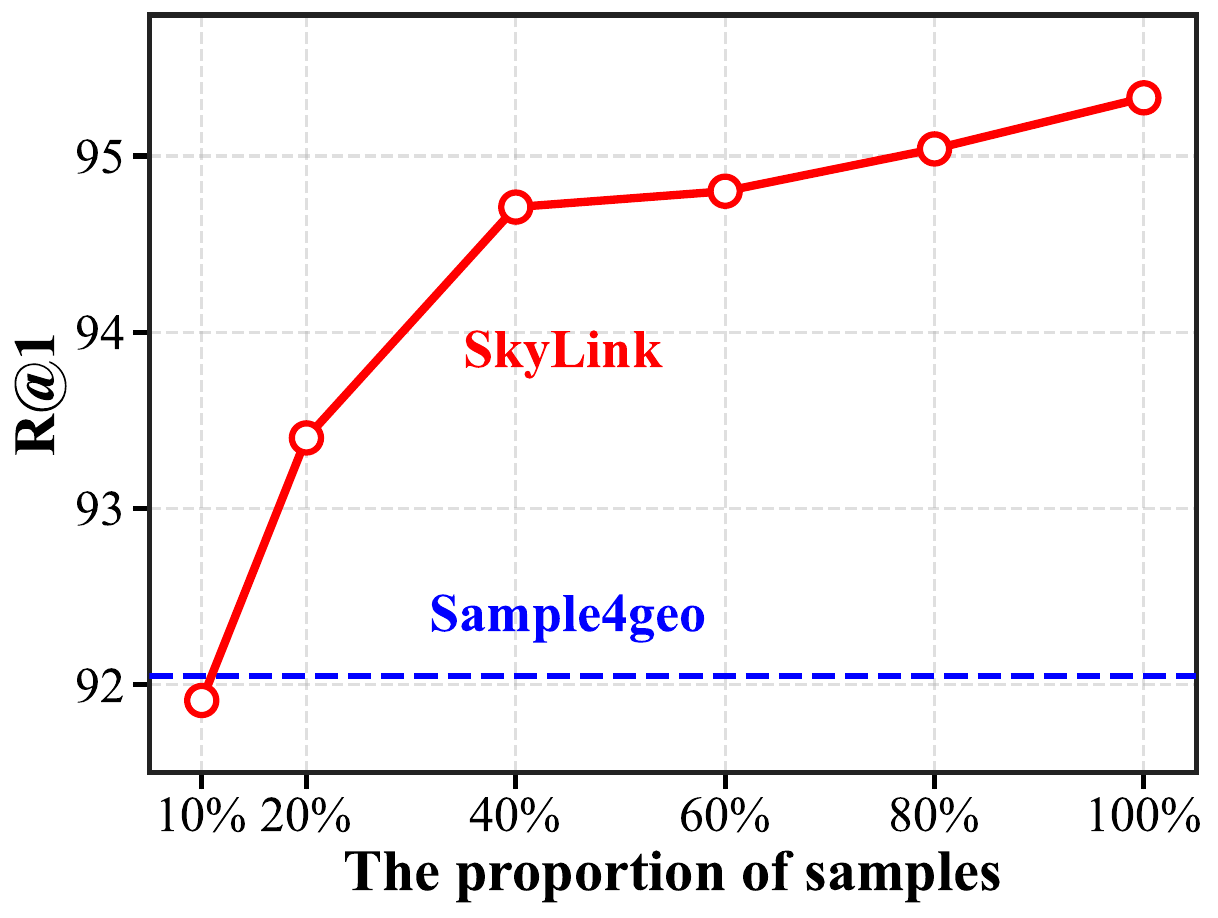}
        \caption{Data Efficiency Analysis}
        \label{0} 
    \end{subfigure}
    \hfill 
    \begin{subfigure}[b]{0.48\textwidth}
        \centering

        \includegraphics[width=0.85\linewidth]{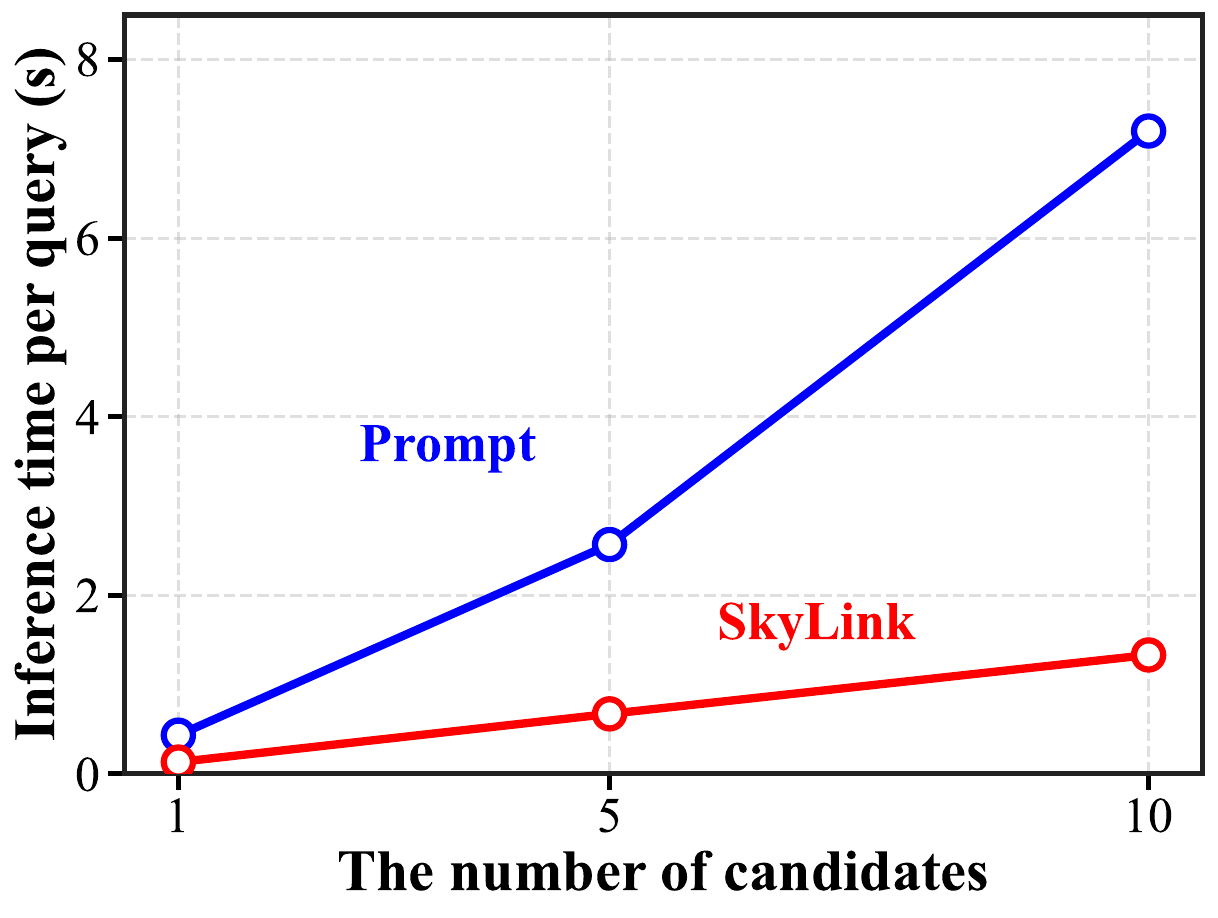}
        \caption{Time Efficiency Analysis}
        \label{time}
    \end{subfigure}
    
    \caption{Efficiency analysis on University-1652.}
    \label{fig:combined_efficiency} 
\end{figure}

In addition to accuracy,  parameter scale and efficiency are also indispensable key indicators in practical deployment scenarios. The study, based on the University-1652 dataset, evaluates the performance of the proposed SkyLink method across three dimensions: parameter scale, time efficiency (measured by inference latency), and data efficiency (assessed by the effectiveness of data utilization). 

In the parameter scale analysis, Table~\ref{2} presents the comparative results of inference accuracy for the SkyLink method using llava-onevision-0.5B \cite{li2024llava}, Qwen2-VL-2B-Instruct \cite{Qwen-VL} and Qwen2-VL-7B-Instruct \cite{Qwen2VL} as backbone models, respectively. The experimental results align with expectations: as the parameter scale of the backbone model increases, the inference accuracy of larger-parameter models consistently outperforms that of smaller-parameter models, indicating that a higher parameter scale contributes to enhancing the model's understanding of task semantics and its capability to capture image features.

In the time efficiency analysis, we adopt a prompt-based method (Appendix~\ref{a11} for details) as the baseline: the query image and all candidate images retrieved by Sample4geo are incorporated into the prompt, which is then used to guide the LVLM to select the most appropriate candidate as the final prediction. As Figure~\ref{time} shows, inference time increases with candidate count (1$\rightarrow$10) for both methods (Skylink incurs computational overhead from additional scoring iterations, whereas the prompting-based baseline suffers from the need to construct progressively longer prompts.), yet Skylink maintains sub-second latency—under one-fifth that of the prompt-based approach. This stems from Skylink’s parallel evaluation architecture, which avoids the near-exponential latency growth of sequential prompt processing. Skylink natively supports large-scale parallel scoring, while the prompt-based method is inherently limited by its sequential input construction, making Skylink highly efficient for large-scale retrieval.

In the data efficiency analysis, Figure~\ref{0} illustrates the performance variation trends of the SkyLink method after fine-tuning with training data of varying scales (with the x-axis representing the proportion of training samples relative to the total sample size, and the y-axis corresponding to the performance metric R@1). The results demonstrate that the accuracy of SkyLink exhibits a stable and continuous upward trend as the training data scale expands, highlighting its superior scalability and generalization capabilities. To further validate data efficiency, the experiments also plot the performance curve of the retriever Sample4geo. Even with fine-tuning using only 20\% of the samples, SkyLink still yields a significant performance increase over the retriever, fully corroborating the high data efficiency advantage of the proposed model, i.e., it requires only limited supervisory signals to effectively enhance the performance of the retriever.

\subsection{Case Study}

\begin{figure}[H]
\centering
\includegraphics[width=0.9\textwidth,scale=0.5]{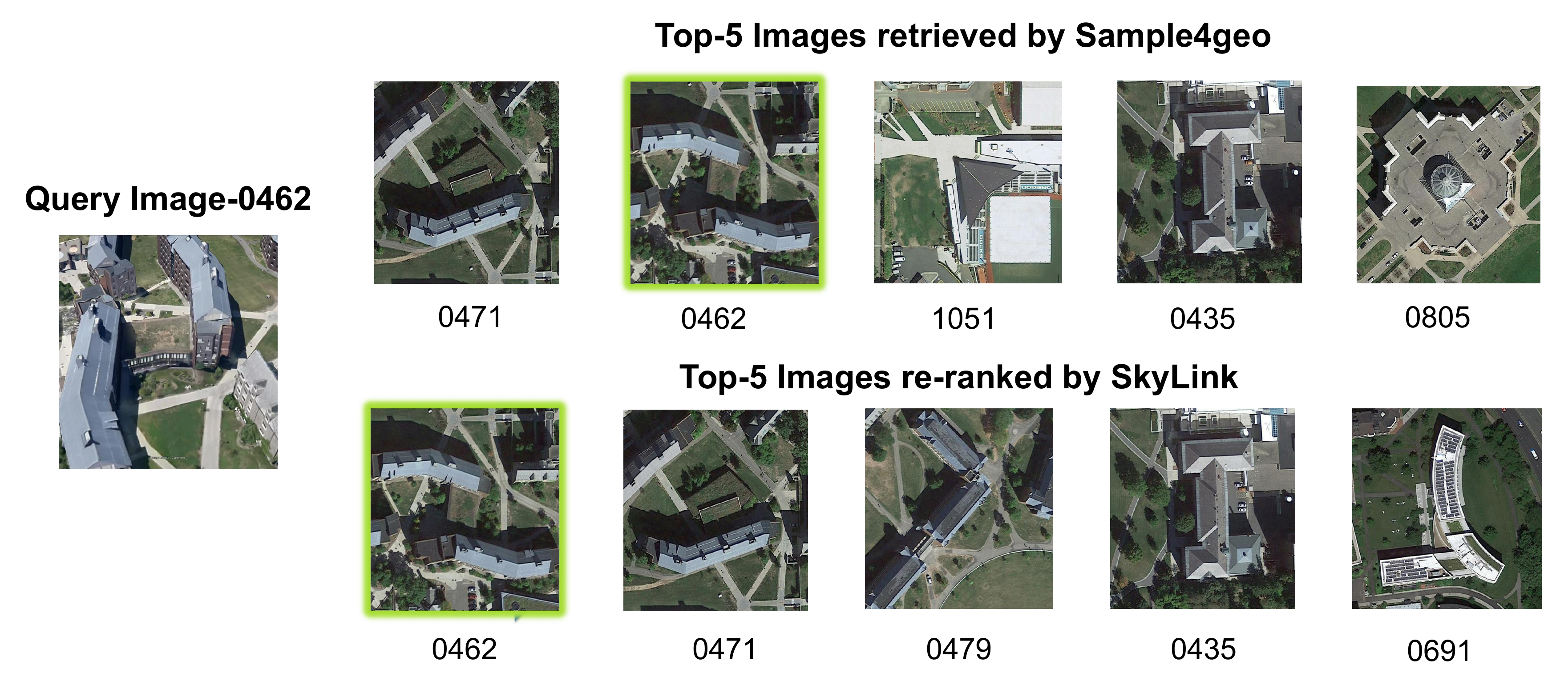}
    \caption{Case Study on the Effectiveness of SkyLink in Candidate Re-Ranking.}
    \label{7}
\end{figure}

To intuitively verify the effectiveness of SkyLink, a qualitative case analysis is presented in Figure~\ref{7}. The upper part of this figure displays the Top-5 retrieval results retrieved by the Sample4geo retrieval model. It can be observed intuitively that while the Sample4geo model retrieves results with relatively high semantic similarity, it still faces significant challenges in handling ambiguous samples. Furthermore, starting from the third retrieved image, the discrepancy between its semantic features and those of the query image increases remarkably, which fails to meet the core requirement of sample relevance for geolocalization tasks. The lower part of the figure presents the results after re-ranking by SkyLink. Experimental results demonstrate that the SkyLink model can not only effectively distinguish between highly similar positive samples and semi-positive samples, but also accurately retrieve satellite images with semantic features similar to those of the query image in terms of architectural style, feature combination (of ground objects), and geometric correlation of roads. This result fully confirms that SkyLink possesses robust geospatial semantic understanding capabilities and efficient re-ranking performance. 

Furthermore, to rigorously validate the effectiveness of the SkyLink similarity-driven soft labeling strategy, we conducted a specialized case study on ambiguous samples within the University-1652 training set. Please refer to Appendix~\ref{a8} for details. 

\section{Related Work}

\subsection{Cross-View UAV Geolocalization}

Cross-view geolocalization is a challenging computer vision task that aims to determine the geographic location of a query image from one perspective (e.g., ground or aerial) by matching it against a database of geo-referenced images from another perspective (typically satellite)~\cite{durgam2024cross,avola2024uav}. Unmanned Aerial Vehicle (UAV) geolocalization~\cite{ji2025game4loc} is a particularly demanding subset of this problem~\cite{10506965,chen2024scale}, characterized by the significant domain gap between oblique-angle UAV imagery and top-down satellite views, which introduces drastic variations in scale, appearance, and geometry. Recent approaches predominantly utilize deep learning to learn viewpoint-invariant feature representations. A prominent research direction focuses on feature partitioning strategies that divide high-level feature maps to capture both fine-grained local details and broader contextual information~\cite{chen2024scale}. MFRGN~\cite{wang2024mfrgn}, employ a dual-flow structure to explicitly model multi-scale global and local information, enhancing generalization to new environments. Alongside feature engineering, advancements in network architectures and training paradigms have been crucial. GeoFormer~\cite{10506965}, for example, uses an efficient Transformer-based Siamese network with multi-scale feature aggregation. Another key area of innovation is the training process itself. Sample4Geo~\cite{Deuser_2023_ICCV} have pioneered the use of hard negative sampling strategies, using either geographical distance or visual similarity to improve the effectiveness of contrastive learning with the InfoNCE loss. While these methods have significantly advanced feature extraction and training strategies, they still largely depend on separate encoders and naive similarity metrics for the final retrieval, which can fail to capture complex cross-view relationships. In contrast, our proposed method leverages a LVLM to jointly model the cross-view interaction between UAV and satellite views, aiming to better capture their complex semantic and spatial relationships for a more accurate final ranking.

\subsection{Large Vision-Language Models}

Large Vision-Language Models (LVLMs)~\cite{zhang2024vision} emerged to address the absence of visual understanding in Large Language Models (LLMs)~\cite{zhao2023survey}, enabling them to process and comprehend multimodal inputs comprising both images and text. A milestone in this progression was the introduction of CLIP (Contrastive Language-Image Pre-training)~\cite{DBLP:conf/icml/RadfordKHRGASAM21} by OpenAI. Through contrastive learning on billions of image-text pairs, CLIP aligns visual and textual features within a shared embedding space. This approach facilitates zero-shot image classification by matching images to corresponding text descriptions, laying the groundwork for subsequent multimodal research. Building on this foundation, a new generation of models such as BLIP~\cite{DBLP:conf/icml/0001LXH22}, LLaVA~\cite{liu2023visual}, Flamingo~\cite{alayrac2022flamingo}, and Qwen-VL~\cite{Qwen2VL} has further advanced the field. The continuous improvements of these LVLMs in cross-modal reasoning and feature representation make them particularly well-suited for complex tasks like cross-view geolocalization.

\subsection{Learning to Rank}

Learning-to-Rank (LTR)~\cite{cao2007learning} is a cornerstone technique in information retrieval~\cite{liu2009learning} and recommender systems~\cite{karatzoglou2013learning}, dedicated to training ranking models that optimize the ordering of candidate items in response to a given query~\cite{kabir2024survey}. LTR methodologies are commonly categorized into three paradigms: pointwise~\cite{bell2018title,dadaneh2020arsm}, pairwise~\cite{tagami2013ctr,cerrato2020fair}, and listwise~\cite{burges2010ranknet,xia2008listwise}. The pointwise approach independently predicts relevance scores for individual query--document pairs; the pairwise paradigm models relative preferences between document pairs; and the listwise framework directly optimizes a global loss function over the entire candidate list. Building upon the LTR framework, this paper incorporates spatial relationships among samples in the UAV geographic information retrieval dataset by introducing a similarity threshold, thereby achieving effective re-ranking.

\section{Conclusion}
In this paper, we propose SkyLink, a plug-and-play LVLM-based re-ranking framework that enhances cross-view UAV geolocalization. To better distinguish ambiguous samples, it employs a dynamic relation-aware loss function with cosine similarity soft labeling for semantic capture. We also introduce SkyRank, the dataset for this re-ranking task. Experiments on University-1652 and SUES-200 show SkyLink boosts retriever performance, outperforming baselines in effectiveness and robustness.

\bibliographystyle{unsrt}  
\bibliography{references}  
\clearpage
 \appendix

\section{Appendix}
\subsection{Datasets and Evaluation Metrics}
\label{a1} 
\begin{table}[H]
  \caption{Details of datasets for experiment.}
  \label{11}
  \begin{center}
    \begin{small}
      \begin{sc}
        \begin{tabular}{lcc}
          \toprule
          Datasets & SUES-200 & University-1652 \\
          \midrule
          Platform          & Drone, Satellite & Drone, Ground, Satellite \\
          Target            & Diversity        & Building \\
          Height difference & True             & False \\
          Training set      & 6120             & 50218 \\
          Images/Location   & 50 + 1           & 54 + 16.64 + 1 \\
          \bottomrule
        \end{tabular}
      \end{sc}
    \end{small}
  \end{center}
  \vskip -0.1in
\end{table}
We conduct experiments on two datasets for cross-view geolocalization tasks: University-1652 and SUES-200, as shown in Table~\ref{11}. Performance is evaluated using Recall and Average Precision (AP) as the assessment metrics.

To quantify the optimization efficacy of the proposed re-ranking strategy in a more scientific and precise manner, the evaluation sample set is partitioned into retrieval sets containing positive samples and those devoid of positive samples when computing AP. For retrieval sets with positive samples, the conventional AP calculation is employed. In contrast, for retrieval sets lacking positive samples, a conservative approach is adopted by assigning an AP value of zero. This methodology yields a lower overall AP compared to the full-set computation, thereby providing a more conservative performance estimate and further underscoring the objectivity and rigor of the re-ranking evaluation process.

\begin{table*}[b]
  \centering
   \caption{Query Images and Corresponding Top 10 Gallery Matches.}
  \label{3}
  \renewcommand{\arraystretch}{1.2} 
  \resizebox{\textwidth}{!}{
  \begin{tabular}{p{3.8cm}|p{11.5cm}}
    \toprule
    \textbf{Query Image} & \textbf{Top 10 Gallery Matches} \\
    \midrule

    \raisebox{-0.5\height}{\includegraphics[width=3.5cm]{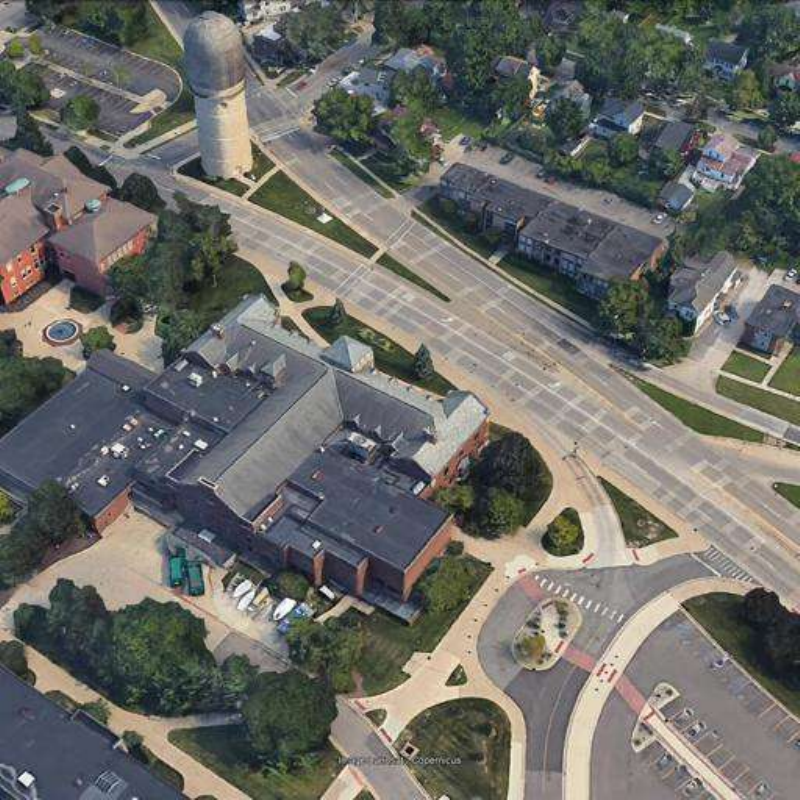}} 
    & 
    \begin{minipage}[t]{11.5cm}
      \centering
      \tikz{
        \node[
          inner sep=2pt,
          draw=green,
          line width=2.5pt,
          rounded corners=3pt,
          drop shadow={shadow xshift=1pt, shadow yshift=-1pt, opacity=0.6}
        ] {\includegraphics[width=0.18\linewidth]{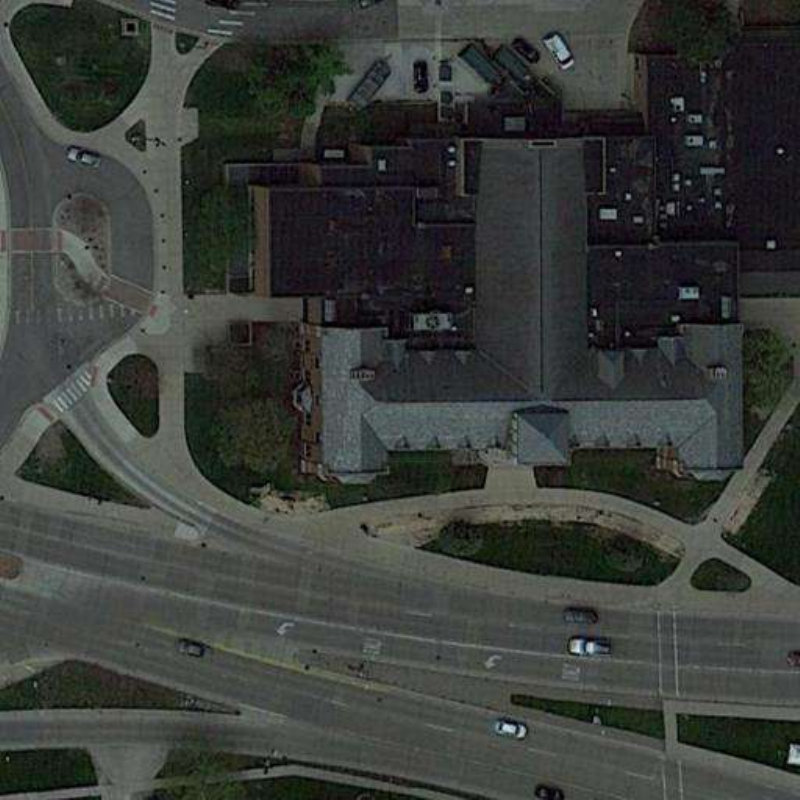}};
      } \hfill
      \includegraphics[width=0.18\linewidth]{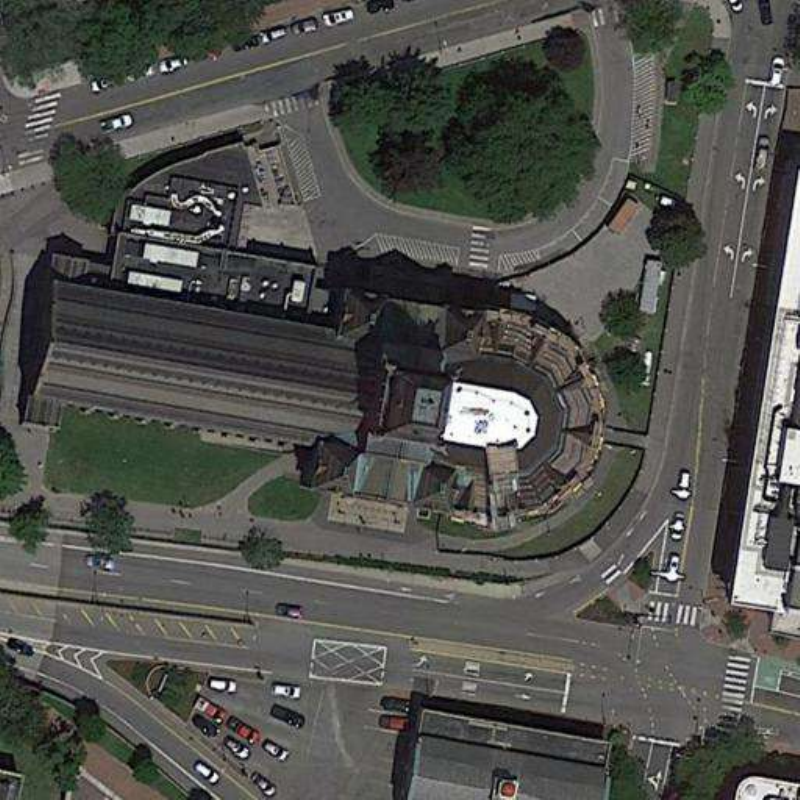} \hfill
      \includegraphics[width=0.18\linewidth]{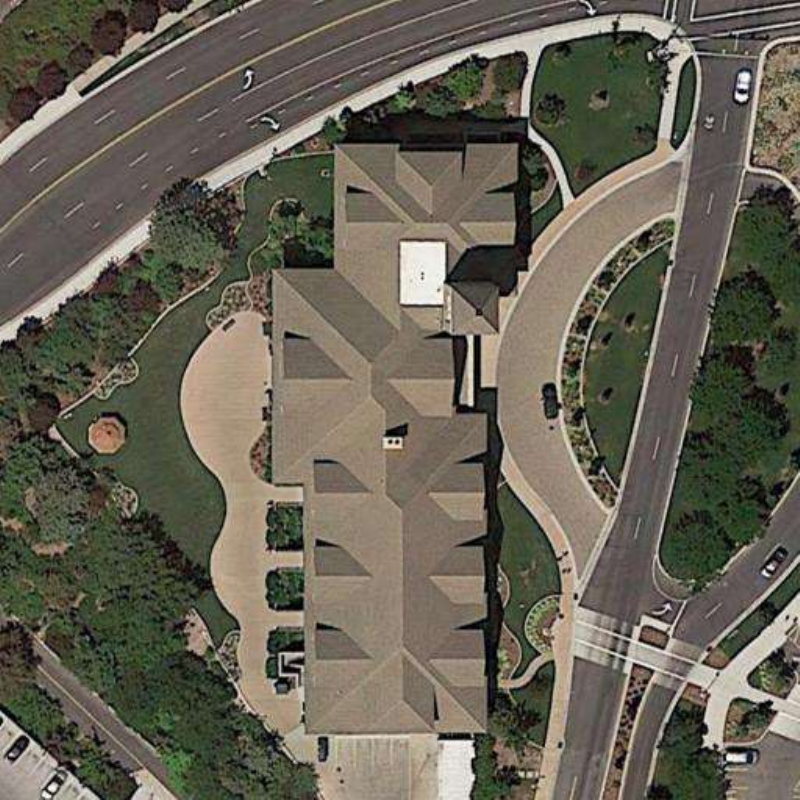} \hfill
      \includegraphics[width=0.18\linewidth]{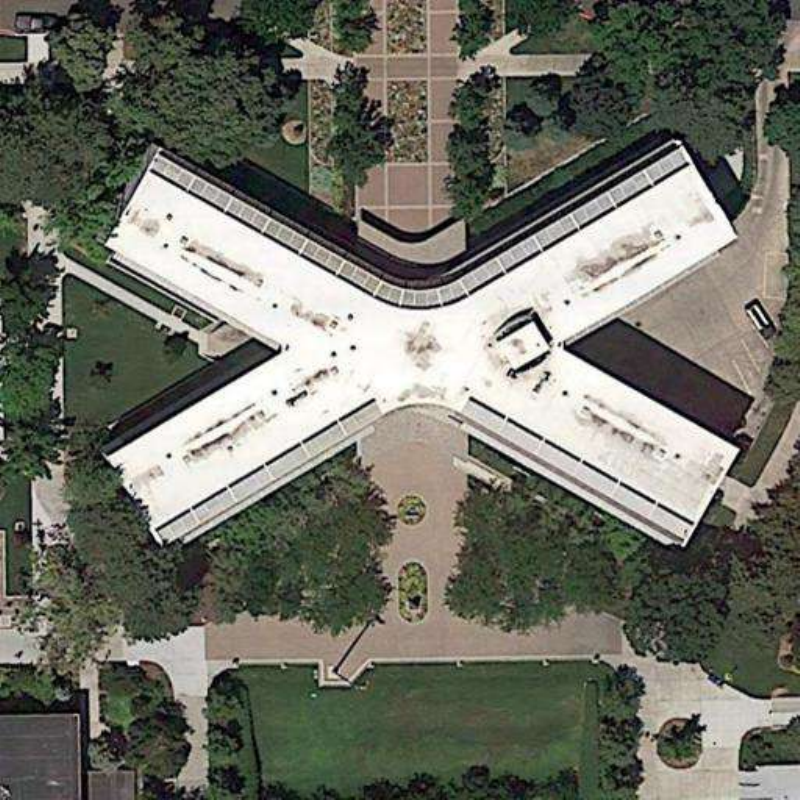} \hfill
      \includegraphics[width=0.18\linewidth]{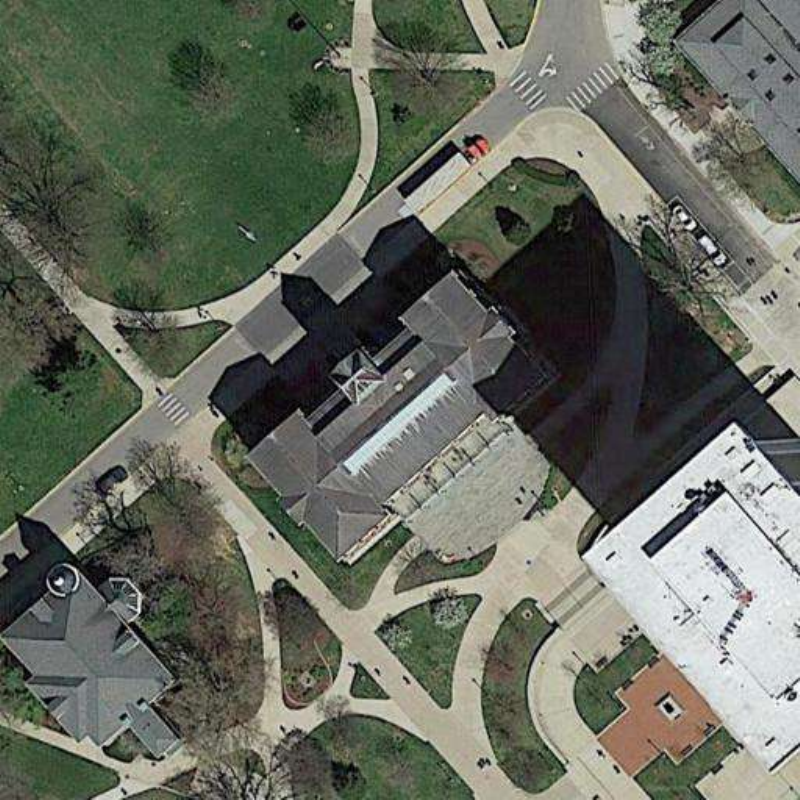} \\ \vspace{3pt}
      \includegraphics[width=0.18\linewidth]{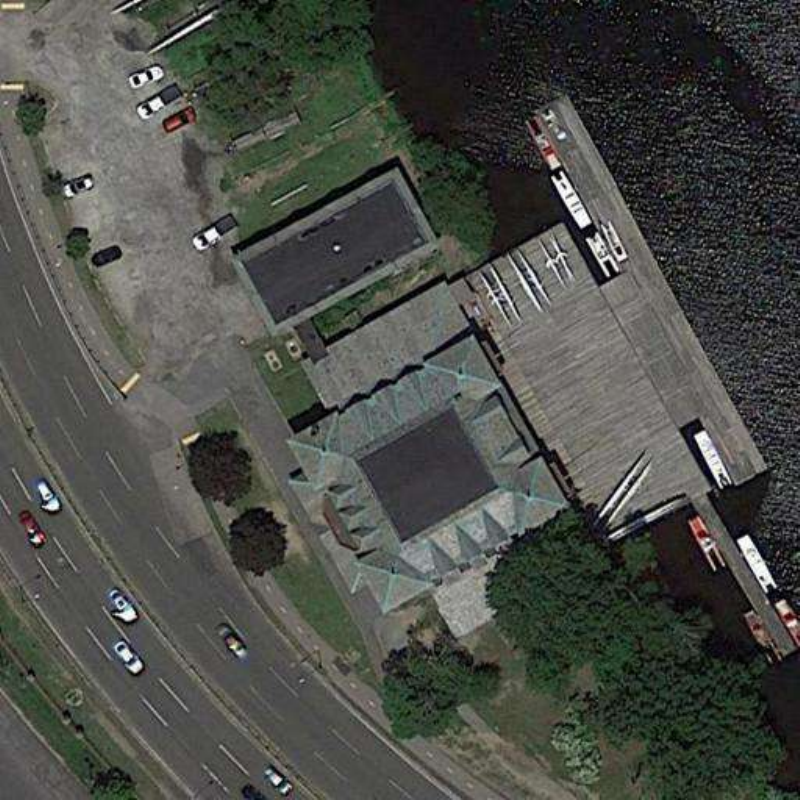} \hfill
      \includegraphics[width=0.18\linewidth]{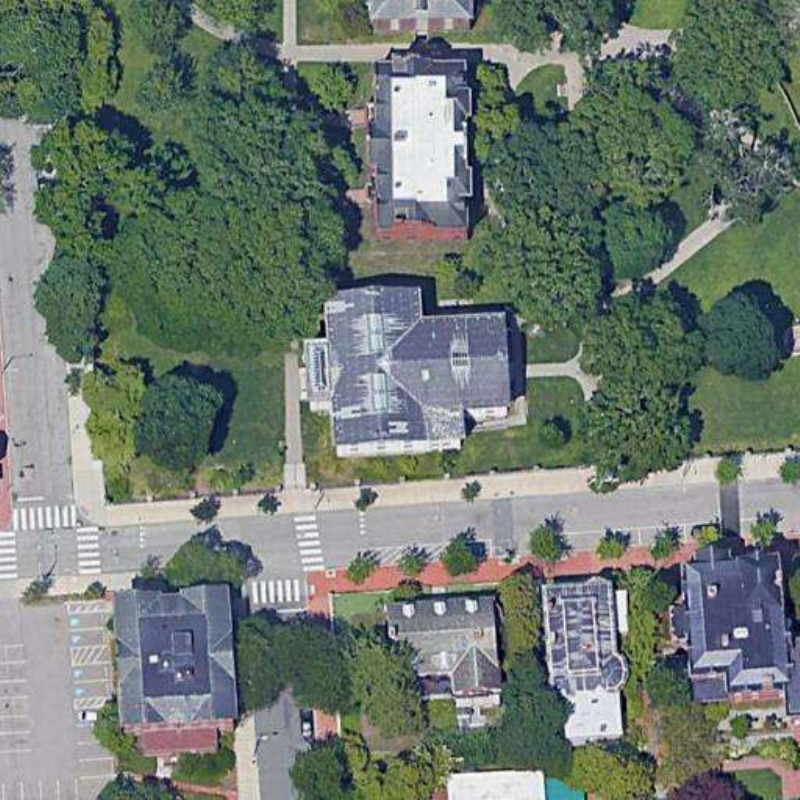} \hfill
      \includegraphics[width=0.18\linewidth]{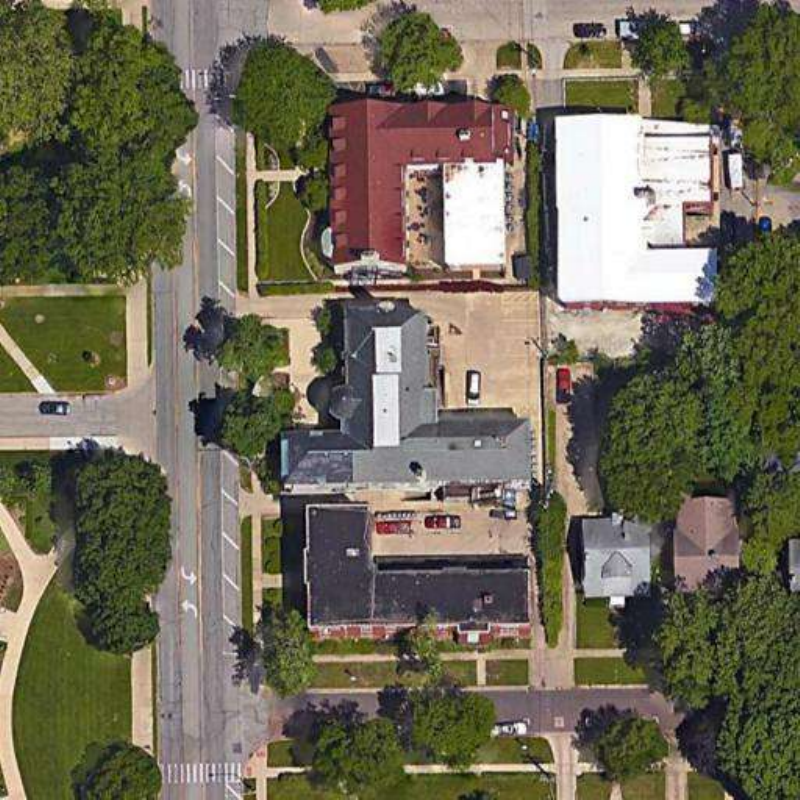} \hfill
      \includegraphics[width=0.18\linewidth]{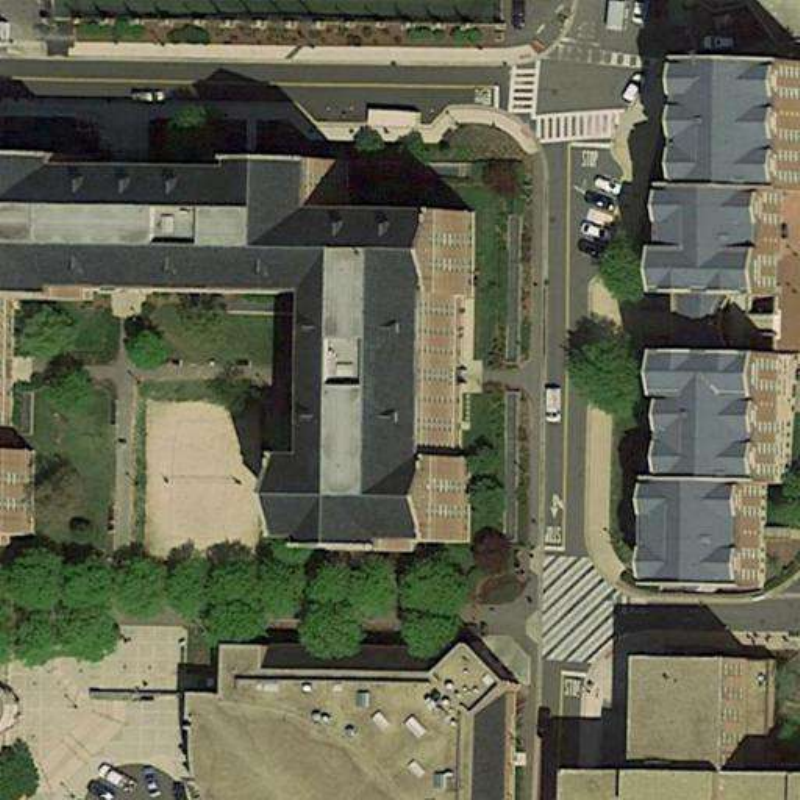} \hfill
      \includegraphics[width=0.18\linewidth]{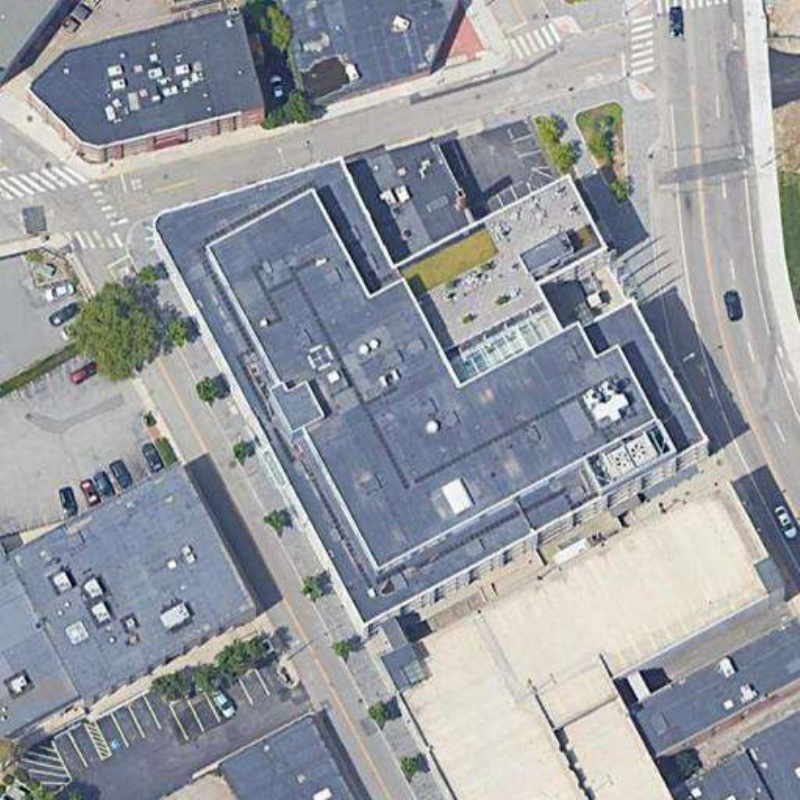}
      
    \end{minipage} \\

    \bottomrule
  \end{tabular}}
  
\end{table*}

\subsection{SkyRank Dataset}
\label{a2} 
This study constructs the first dataset for training candidate ranking models in cross-view geolocalization, with a total of 5 sub-datasets built to cover the data distributions of University-1652 and SUES-200, as illustrated in Table~\ref{3}.

\subsection{The Introduction of Retrievers}
\label{a3} 

SDPL introduces a shifting-dense partition strategy that divides images into dense parts while preserving global structure. A shifting-fusion mechanism generates multiple part sets from varied centers and adaptively fuses features, enhancing robustness to target shifts and scale variations.\\

Sample4Geo adopts a compact contrastive pipeline with symmetric InfoNCE loss, bypassing aggregation modules and preprocessing. It introduces dual hard-negative mining via geographic proximity and embedding similarity, enhancing feature discrimination. \\

MCCG builds on ConvNeXt, using cross-dimensional interaction to yield multiple discriminative representations. This design captures extensive contextual details, overcoming information loss in global or segmented features.

\subsection{More Details on Training and Inference}
\label{a4} 
Our experiments are divided into two phases: training and inference. The training phase experiments were conducted on 2 NVIDIA A100 GPUs, while the inference phase was performed on 1 NVIDIA A100 GPU. The results indicate that in the University-1652, each training epoch requires approximately 13 hours, while in the SUES-200 single-altitude level, each training epoch takes about 4 hours, with a single-device GPU memory consumption of approximately 60 GB in both cases. (Refer to Table~\ref{10})

\begin{table}[t]
  \caption{Parameter setting.}
  \label{10}
  \begin{center}
    \begin{small}
      \begin{sc}
        \setlength{\tabcolsep}{4pt} 
        \begin{tabular}{ll}
          \toprule
          Parameter & Value \\
          \midrule
          Training Time (Univ-1652) & 13 hours / epoch \\
          Training Time (SUES-200)  & 4 hours / epoch  \\
          VLM model                 & Qwen2-VL-7B \\
          Deepspeed                 & Stage 2 \\
          GPU Memory Consumption    & 30 GB / GPU \\
          Batch Size per Device     & 1 \\
          LoRA rank                 & 16 \\
          Scaling factor            & 32 \\
          LoRA dropout              & 0.05 \\
          Learning rate             & $10^{-4}$ \\
          Batch size                & 4 \\
          Training epoch            & 1 \\
          \bottomrule
        \end{tabular}
      \end{sc}
    \end{small}
  \end{center}
  \vskip -0.1in
\end{table}

\subsection{More Details on Hyperparameter Analysis}
\label{a5} 
Extensive experiments were conducted on the SUES-200 dataset to validate the model performance. Experimental results demonstrate that when the hyperparameter $T$ approaches 1, the rate of performance improvement of the model slows down in the challenging task of low-altitude UAV image retrieval from satellite imagery. In contrast, for the high-altitude target retrieval task, the heterogeneity of the reference image set shows an upward trend, while the requirement for processing ambiguous samples during the training phase decreases accordingly. Ultimately, this leads to a more significant performance improvement of the model in this specific task scenario. (Refer to Table~\ref{4})

Furthermore, through sensitivity analysis, we identified a set of optimal hyperparameter configurations that can maximize the re-ranking potential of the large model (\( |G_z| \) = 13, \( |\text{top-k}| = 10 \),  \( T = 0.9 \)).

\begin{table}[t]
  \caption{Sensitivity analysis of hyperparameter $T$ on the SUES-200 dataset.}
  \label{4}
  \begin{center}
    \begin{small}
      \begin{sc}
        
        \setlength{\tabcolsep}{3.5pt} 
        \begin{tabular}{lcccccccc}
          \toprule
          & \multicolumn{2}{c}{SUES-150} & \multicolumn{2}{c}{SUES-200} & \multicolumn{2}{c}{SUES-250} & \multicolumn{2}{c}{SUES-300} \\
          \cmidrule(lr){2-3} \cmidrule(lr){4-5} \cmidrule(lr){6-7} \cmidrule(lr){8-9}
          Method & R@1$\uparrow$ & AP$\uparrow$ & R@1$\uparrow$ & AP$\uparrow$ & R@1$\uparrow$ & AP$\uparrow$ & R@1$\uparrow$ & AP$\uparrow$ \\
          \midrule
          Sample4geo          & 92.87 & 95.58 & 94.47 & 96.79 & 96.83 & 98.17 & 97.40 & 98.16 \\
          +SkyLink ($T=0.9$)  & \textbf{95.77} & \textbf{97.51} & \textbf{96.45} & \textbf{98.02} & 97.75 & 98.70 & 97.62 & 98.76 \\
          +SkyLink ($T=1.0$)  & 95.20 & 97.10 & 95.95 & 97.84 & \textbf{97.98} & \textbf{98.93} & \textbf{98.92} & \textbf{99.41} \\
          \bottomrule
        \end{tabular}
      \end{sc}
    \end{small}
  \end{center}
  \vskip -0.1in
\end{table}

\subsection{The Supplement of Ablation Study}
\label{a6}

\begin{figure}[H]
\centering

\begin{subfigure}[b]{\linewidth}
    \centering
    \includegraphics[width=\linewidth]{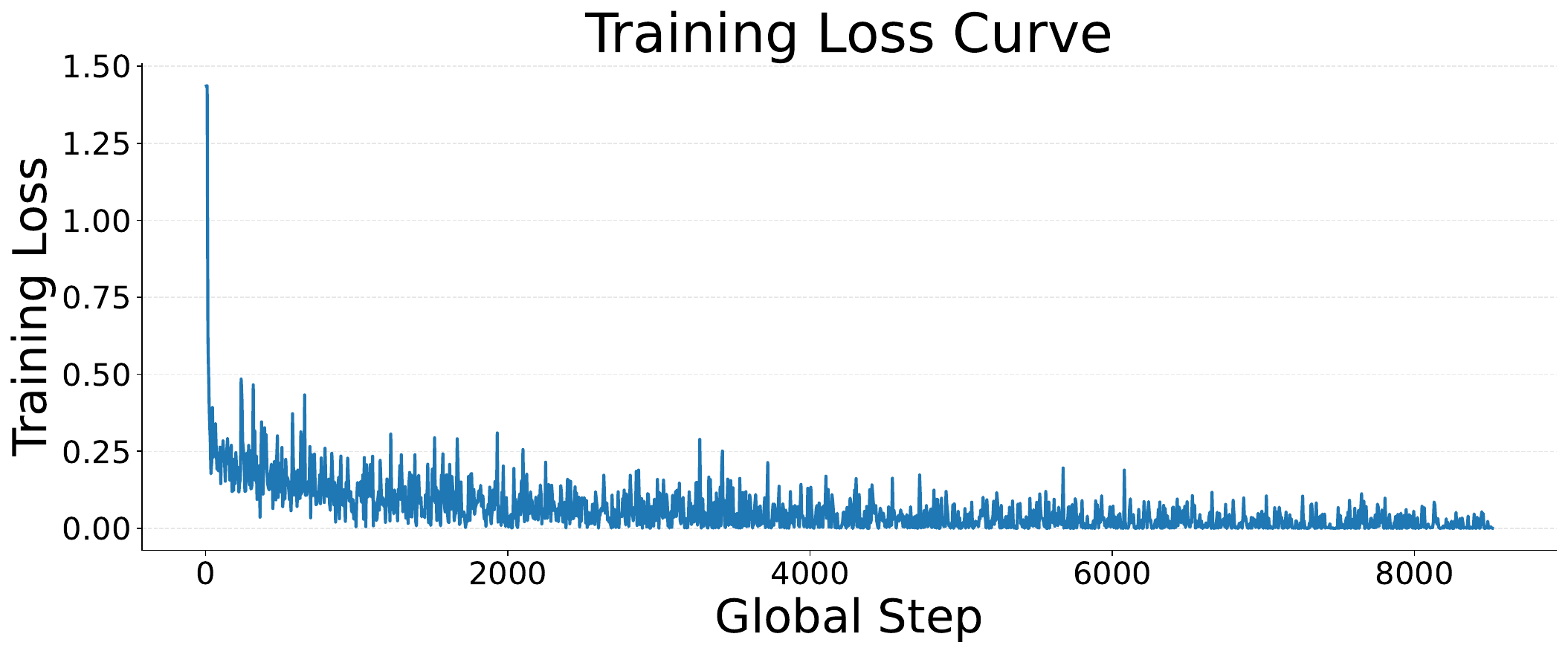}
    \caption{Training loss of the model (with similarity threshold)}
    \label{figure.loss_with_threshold}
\end{subfigure}

\vspace{1em} 

\begin{subfigure}[b]{\linewidth}
    \centering
    \includegraphics[width=\linewidth]{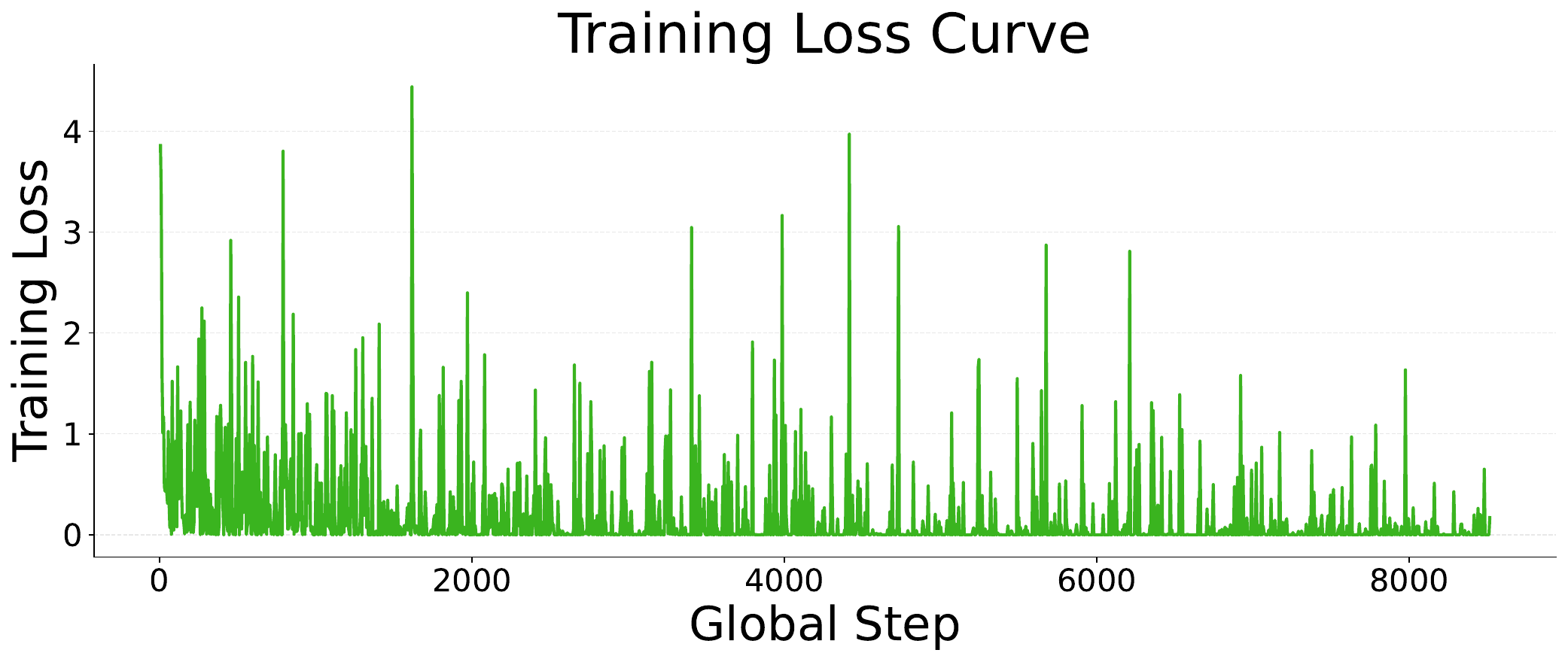}
    \caption{Training loss of the model (without similarity threshold)}
    \label{figure.loss_without_threshold}
\end{subfigure}

\caption{Comparison of training loss with and without similarity threshold}
\label{8}
\end{figure}

After removing the similarity threshold mechanism from the model (i.e., w/o similarity threshold), we observed severe fluctuations in the loss value throughout the model training process via the experimental training monitoring platform W\&B. (Figure~\ref{8}) This result reveals that simply using the first reference sample directly as the label for training (i.e., without screening and optimizing labels through the similarity threshold) will have a devastating impact on the model's ability to process similar samples.

\subsection{Case Study on Soft Labels}
\label{a8} 
\begin{figure}[H]
\centering
\hspace*{-13pt}
\includegraphics[width=\textwidth,scale=0.5]{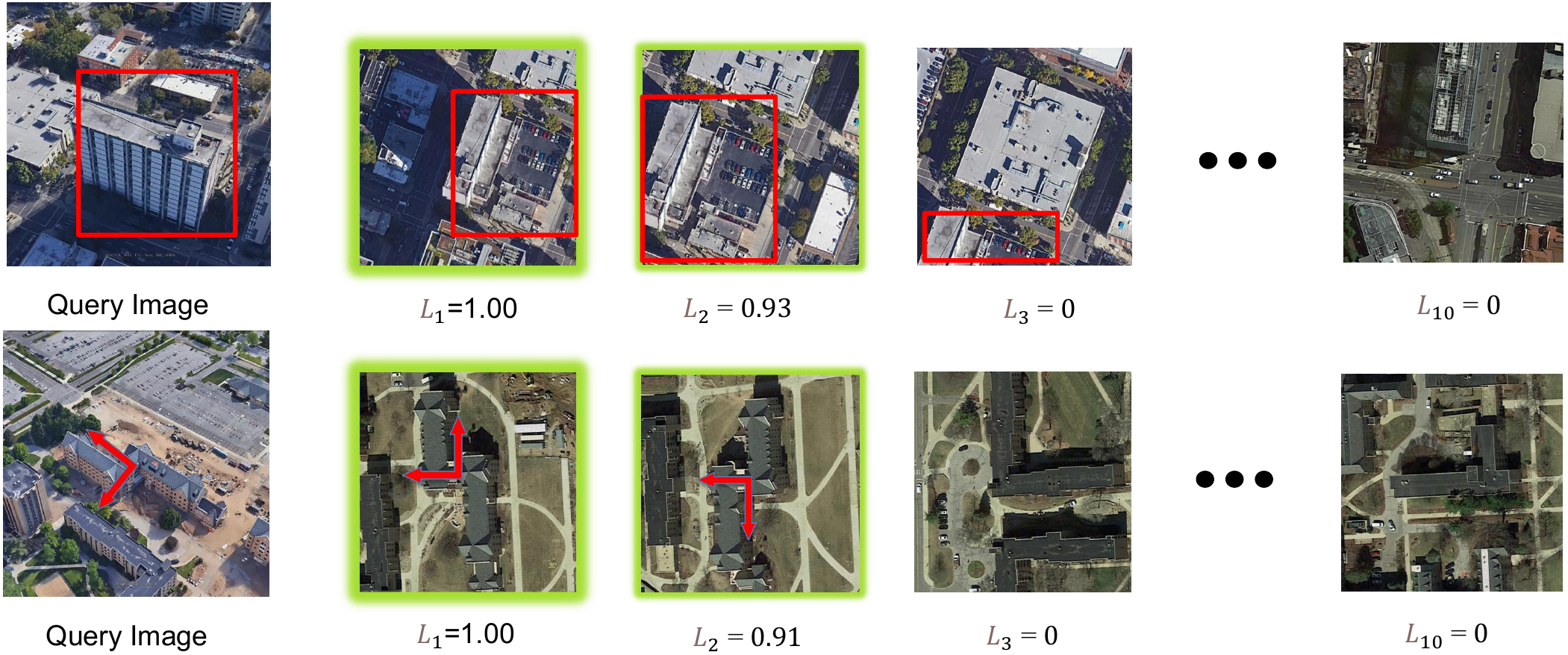}
    \caption{Case Study on the similarity labels of SkyLink.}
    \label{6}
\end{figure}

To rigorously evaluate the efficacy of SkyLink's similarity-based soft labeling mechanism, we conducted a targeted case study on ambiguous samples from the University-1652 training set, as illustrated in Figure~\ref{6}. For a given query image, the original hard label designates only the top-ranked reference image as the positive sample, thereby overlooking the supervisory potential of visually similar yet ambiguous instances. These ambiguous samples typically correspond to different orientations or viewpoints of the same geographic region. By employing soft labeling, additional discriminative signals can be effectively harnessed from such instances. Specifically, with the similarity threshold set at 0.9, samples exceeding this threshold are assigned positive soft labels proportional to their cosine similarity scores, whereas those below the threshold receive zero labels. This mechanism enables SkyLink to reliably propagate pseudo-labels to visually akin but label-ambiguous drone-view and satellite-view instances, thereby substantially mitigating label noise and enhancing cross-view geo-localization performance.

\subsection{The prompt of prompt-based method}
\label{a11}

In our baseline, the large vision-language model (VLM) is prompted with a multi-image context that includes one query image and up to top-k candidate reference images. The model is instructed to assign a relevance score between 0 and 10 for each candidate relative to the query. The core prompt template can be abstracted as:

\begin{tcolorbox}[colback=gray!5!white,colframe=gray!75!black]
\textcolor{myblue}{\{query image\}} \\
This is the query image above. Now, here are \{k\} candidate reference images. For each candidate, evaluate how well it matches the query image on a scale from 0 (no match) to 10 (perfect match). \\
\textcolor{myblue}{\{candidate image 1\}} \\
\textcolor{myblue}{\{candidate image 2\}} \\
$\vdots$ \\
\textcolor{myblue}{\{candidate image k\}} \\
Output only the scores in this exact format: \\
Candidate 1: \{score\} \\
Candidate 2: \{score\} \\
$\vdots$ \\
Candidate k: \{score\} \\
Do not add any other text or explanation.
\end{tcolorbox}

\subsection{Algorithm}
\label{a10}

SkyLink is a retrieval-augmented generative framework that aligns visual-semantic representations through contrastive supervision derived from a vision-language model (VLM). During training, for each query image \(I_q\), SkyLink constructs a context set \(\mathcal{K}\) comprising the ground-truth target \(I_{gt}\) and \(k-1\) top candidates from a coarse retrieval pool \(\mathcal{C}'_q\). A lightweight LoRA-adapted large vision-language model (\(\mathcal{M}_{\text{LoRA}}\)) processes multimodal prompts formed by pairing \(I_q\) with each candidate in \(\mathcal{K}\), producing last-layer embeddings that are scored via a learnable linear head \(w\). These scores are supervised by soft labels \(L_j\) computed as thresholded cosine similarities between retrieved and ground-truth image embeddings from a frozen dual-encoder retriever \((E_q, E_r)\), enabling gradient-based optimization of both LoRA parameters and the scoring head. At inference, SkyLink first retrieves the top-\(m\) candidates from a large reference corpus \(\mathcal{R}\) using \(E_q\), then re-ranks them by prompting \(\mathcal{M}_{\text{SkyLink}}\) with each candidate and selecting the one yielding the highest score.

\begin{algorithm}[H]
\caption{SkyLink: Training and Inference}
\label{alg:skylink}
\begin{algorithmic}[1]

\Require 
\Statex \quad -- Training: $\mathcal{D}_{\text{train}} = \{ (I_q, \mathcal{C}'_q, I_{gt}) \}$, retriever $(E_q, E_r)$, LVLM $\mathcal{M}$, LoRA rank $r$, $k$, threshold $T$
\Statex \quad -- Inference: $I_q$, $\mathcal{R} = \{I_r^i\}_{i=1}^N$, $V_R = \{E_r(I_r^i)\}$, $\mathcal{M}_{\text{SkyLink}}$, $m$
\Ensure $\hat{I}$

\Statex \hrulefill \quad \textbf{Training Stage} \quad \hrulefill
\Procedure{TrainSkyLink}{$\mathcal{D}_{\text{train}}, \mathcal{M}, E_r, k, T, r$}
    \State Initialize LoRA $\{A_l, B_l\}_{l=1}^L$ in $\mathcal{M}$
    \For{each $(I_q, \mathcal{C}'_q, I_{gt}) \in \mathcal{D}_{\text{train}}$}
        \State $\mathcal{K} \gets \{I_{gt}\} \cup \text{top-}(k-1)(\mathcal{C}'_q)$ \Comment{$|\mathcal{K}|=k$}
        \State $\mathcal{P} \gets \{ \text{Prompt}(I_q, I_c) \mid I_c \in \mathcal{K} \}$
        \State $h[\text{last}] \gets \mathcal{M}_{\text{LoRA}}(\mathcal{P})[-1] \in \mathbb{R}^{k \times d}$
        \State $S \gets w^\top h[\text{last}] \in \mathbb{R}^k$
        \State $L_j \gets \max\left(0, \frac{E_r(I_{c_j})^\top E_r(I_{gt})}{\|E_r(I_{c_j})\| \|E_r(I_{gt})\|} - T \right)$ for $j=1$ to $k$ \Comment{Soft labels via cosine sim}
        \State $\mathcal{L} \gets -\frac{1}{k}\sum_{j=1}^k \bigl[ L_j \log\sigma(S_j) + (1-L_j) \log(1-\sigma(S_j)) \bigr]$
        \State Update LoRA $\{A_l,B_l\}$ and $w$ via $\nabla\mathcal{L}$
    \EndFor
    \State \Return $\mathcal{M}_{\text{SkyLink}}$
\EndProcedure

\Statex \hrulefill \quad \textbf{Inference Stage} \quad \hrulefill
\Procedure{InferSkyLink}{$I_q$, $\mathcal{R}$, $V_R$, $\mathcal{M}_{\text{SkyLink}}$, $m$}
    \State $v_q \gets E_q(I_q)$
    \State $C_q \gets \text{top-$m$}(V_R, v_q)$ \Comment{Retrieve via cosine similarity}
    \State $s \gets \mathcal{M}_{\text{SkyLink}}(\text{Prompt}(I_q, I_c))$ for each $I_c \in C_q$
    \State \Return $\hat{I} \gets \arg\max_{I_c \in C_q} s[I_c]$
\EndProcedure

\end{algorithmic}
\end{algorithm}

\subsection{Limitation}
\label{a9}
SkyLink achieves significant improvements in the accuracy and robustness of cross-view UAV geolocalization, particularly through its dynamic relation-aware loss function and the SkyRank dataset. However, the method is not without limitations. The plug-and-play framework adopted by SkyLink introduces an additional re-ranking stage, which significantly increases computational overhead during inference compared to direct embedding-based retrieval methods. Currently, this added complexity constrains its deployment on airborne platforms, often necessitating reliance on ground stations for processing.  This poses substantial practical challenges in resource-constrained environments or applications with stringent real-time requirements. Future work could explore the development of end-to-end ranking approaches to mitigate this issue. Furthermore, as a re-ranking mechanism, SkyLink's performance is strictly bounded by the recall of the initial retrieval stage. The model can only re-rank candidates provided by the base retriever. If the ground truth is not present within the top-$N$ candidates (where $N=10$ in our inference setting), SkyLink cannot recover the correct match. Consequently, the theoretical upper bound of our method is determined by the Retriever's R@$N$. Future efforts should focus on developing more advanced retrievers to improve this initial recall ceiling or expanding the candidate pool size $N$, albeit at the cost of increased computational load.

\end{document}